\documentclass[a4paper,fleqn]{cas-sc}
\usepackage[numbers,square]{natbib}
\usepackage{multicol}
\usepackage{multirow}
\usepackage{subfig}
\usepackage{graphicx}
\usepackage{float}
\usepackage{appendix}
\usepackage{soul}

\def\tsc#1{\csdef{#1}{\textsc{\lowercase{#1}}\xspace}}
\tsc{WGM}
\tsc{QE}

\begin{document}
\let\WriteBookmarks\relax
\def\floatpagepagefraction{1}
\def\textpagefraction{.001}

\shorttitle{Analysis and Applications of Deep Learning with Finite Samples in Full Life-Cycle Intelligence of NPG}  

\shortauthors{Tang et al. (2023)}

\title [mode = title]{Analysis and Applications of Deep Learning with Finite Samples in Full Life-Cycle Intelligence of Nuclear Power Generation}  

\author[a1,a2]{Chenwei Tang}

\author[a1,a2]{Wenqiang Zhou}

\author[a3]{Dong Wang}
%\ead{wangdongby@outlook.com}

\author[a1,a2]{Caiyang Yu}

\author[a1,a2]{Jiancheng Lv}
%\ead{lvjiancheng@scu.edu.cn}

\author[a1,a2]{Zhenan He}
%\ead{zhenan@scu.edu.cn}

\author[a1,a2]{Jizhe Zhou}

\author[a1,a2]{Shudong Huang}

\author[a1,a2]{Yi Gao}

\author[a4]{Jianming Chen}
				
\author[a1,a2]{Wentao Feng}\corref{cor1}
\ead{Wtfeng2021@scu.edu.cn}
\cortext[cor1]{Corresponding author}

\affiliation[a1]{organization={College of Computer Science},
            addressline={Sichuan University}, 
            city={Chengdu},
            postcode={610065}, 
            country={China}}
            
\affiliation[a2]{organization={Engineering Research Center of Machine Learning and Industry Intelligence},
            addressline={Ministry of Education}, 
            city={Chengdu},
            postcode={610065}, 
            country={China}}

\affiliation[a3]{organization={Science and Technology on Reactor Fuel and Materials Laboratory},
            addressline={Nuclear Power Institute of China}, 
            city={Chengdu},
            postcode={610213}, 
            country={China}}

\affiliation[a4]{organization={Institute of Software},
            addressline={Chinese Academy of Sciences}, 
            city={Beijing},
            postcode={100190}, 
            country={China}}

\begin{abstract}
The advent of Industry 4.0 has precipitated the incorporation of Artificial Intelligence (AI) methods within industrial contexts, aiming to realize intelligent manufacturing, operation as well as maintenance, also known as industrial intelligence. However, intricate industrial milieus, particularly those relating to energy exploration and production, frequently encompass data characterized by long-tailed class distribution, sample imbalance, and domain shift. These attributes pose noteworthy challenges to data-centric Deep Learning (DL) techniques, crucial for the realization of industrial intelligence. The present study centers on the intricate and distinctive industrial scenarios of Nuclear Power Generation (NPG), meticulously scrutinizing the application of DL techniques under the constraints of finite data samples. Initially, the paper expounds on potential employment scenarios for AI across the full life-cycle of NPG. Subsequently, we delve into an evaluative exposition of DL's advancement, grounded in the finite sample perspective. This encompasses aspects such as small-sample learning, few-shot learning, zero-shot learning, and open-set recognition, also referring to the unique data characteristics of NPG. The paper then proceeds to present two specific case studies. The first revolves around the automatic recognition of zirconium alloy metallography, while the second pertains to open-set recognition for signal diagnosis of machinery sensors. These cases, spanning the entirety of NPG's life-cycle, are accompanied by constructive outcomes and insightful deliberations. By exploring and applying DL methodologies within the constraints of finite sample availability, this paper not only furnishes a robust technical foundation but also introduces a fresh perspective toward the secure and efficient advancement and exploitation of this advanced energy source.
\end{abstract}

\begin{highlights}
\item Identification of the challenges faced by data-driven Deep Learning (DL) methods in intricate nuclear power generation (NPG) scenarios.
\item Review of the progress of DL technologies from the finite sample perspective, which is relevant to addressing the challenges faced in industrial settings of NPG.
\item Presentation of two case studies demonstrating the applicability of DL methods with finite samples in the full life-cycle intelligence of NPG.
\item Providing a technical basis and novel perspective for the safer and more efficient development and utilization of nuclear energy.
\end{highlights}


\maketitle

\section{Introduction}
\label{intro}

\subsection{Backgroud}
\label{background}
Over the last century, global energy consumption has steadily increased, driven by factors such as population growth, expanding economies, and increased industrial productivity. Despite scientific and technological efforts to improve the efficiency of fossil fuel usage to meet rising demand, the replacement of fossil fuels with cleaner alternatives is becoming increasingly inevitable in the coming decades \cite{longden2022clean}. This is due to the non-renewable nature of fossil fuels, their limited reserves, and their detrimental environmental impact. However, current renewable energy sources such as wind, solar, and hydro are not yet capable of fully replacing fossil fuels \cite{al2023analysis}. This is because they are highly dependent on natural conditions, such as weather and tides, resulting in intermittent, stochastic, and volatile power generation. To address these issues, many countries are undergoing an energy transition that involves developing smarter and more robust energy transmission grids, increasing storage capacities, and implementing cleaner and more stable backup energy sources \cite{hong2014nuclear}. In this context, the development of safer and more intelligent Nuclear Power Generation (NPG) technologies has become an essential option to facilitate the transformation from fossil fuels to renewable energy systems and provide reliable backup energy \cite{nian2014life}.

Since the discovery of nuclear fission in the 1930s and its subsequent development for civilian use, the efficient conversion of nuclear energy into power has become a sophisticated technology with high-tech and strategic attributes since its inception \cite{hiebert1988role}. It involves complex scientific and engineering issues, ranging from the exploration and mining of radioactive minerals such as uranium-235, the preparation of nuclear fuel rods, and the construction of nuclear power plants throughout the full life-cycle of NPG. Although its principle of power generation still relies on the fission reaction of nuclear fuel in a reactor, which produces a large amount of heat to generate high-temperature and high-pressure steam to drive turbines, nuclear power is still considered a stable and reliable clean energy source due to its long-lasting energy release and the significant reduction of hazardous waste after reprocessing,  harmless disposal or deep burial. As of Sep. 2022, there are 437 operational civilian nuclear reactors worldwide, with a total installed capacity of 393 GW, distributed across 32 different countries \cite{brunnengraber2015nuclear}. Among them, most of the technically advanced third-generation reactors are located in Asia.

As of today, with the gradual maturity of the nuclear industry and the rapid development of Artificial Intelligence (AI), the demand for intelligentization in the field of nuclear power has become increasingly urgent \cite{lyu2021artificial}. The application of advanced technologies, such as Deep Learning (DL), at various stages of the nuclear power life cycle significantly improves industrial production quality and efficiency, and effectively reduces operating costs \cite{shi2021machine}. Additionally, NPG involves radioactive waste, and reactor malfunctions may lead to radioactive material leakage. In this context, AI assists in the post-processing of depleted fuel and the processing and disposal of radioactive waste while also making daily maintenance and accident rescue at nuclear power plants more intelligent. It will greatly reduce the risks to human operators while ensuring the safe operation of nuclear power plants. Moreover, the rapid development of AI technology has facilitated its deep integration with other techniques in the industry, particularly in the energy and manufacturing industries \cite{tang2022deep}. However, in the face of some practical requirements and objective conditions in the industrial field, especially in scenarios such as anomaly detection/fault diagnosis, where people do not want sample events to occur but still require higher work efficiency from AI tools, DL technologies with finite samples, including Small-Sample Learning (SSL) \cite{liu2019network}, Few-Shot Learning (FSL) \cite{wang2020generalizing}, Zero-Shot Learning (ZSL) \cite{AWA2-gbu}, and Open-Set Recognition (OSR) \cite{geng2020recent}, need to be applied. These techniques will enable AI tools to work with limited training samples and still achieve high accuracy and efficiency.

\begin{figure*}[t]%[htbp]
\centering
\includegraphics[width=0.93\textwidth]{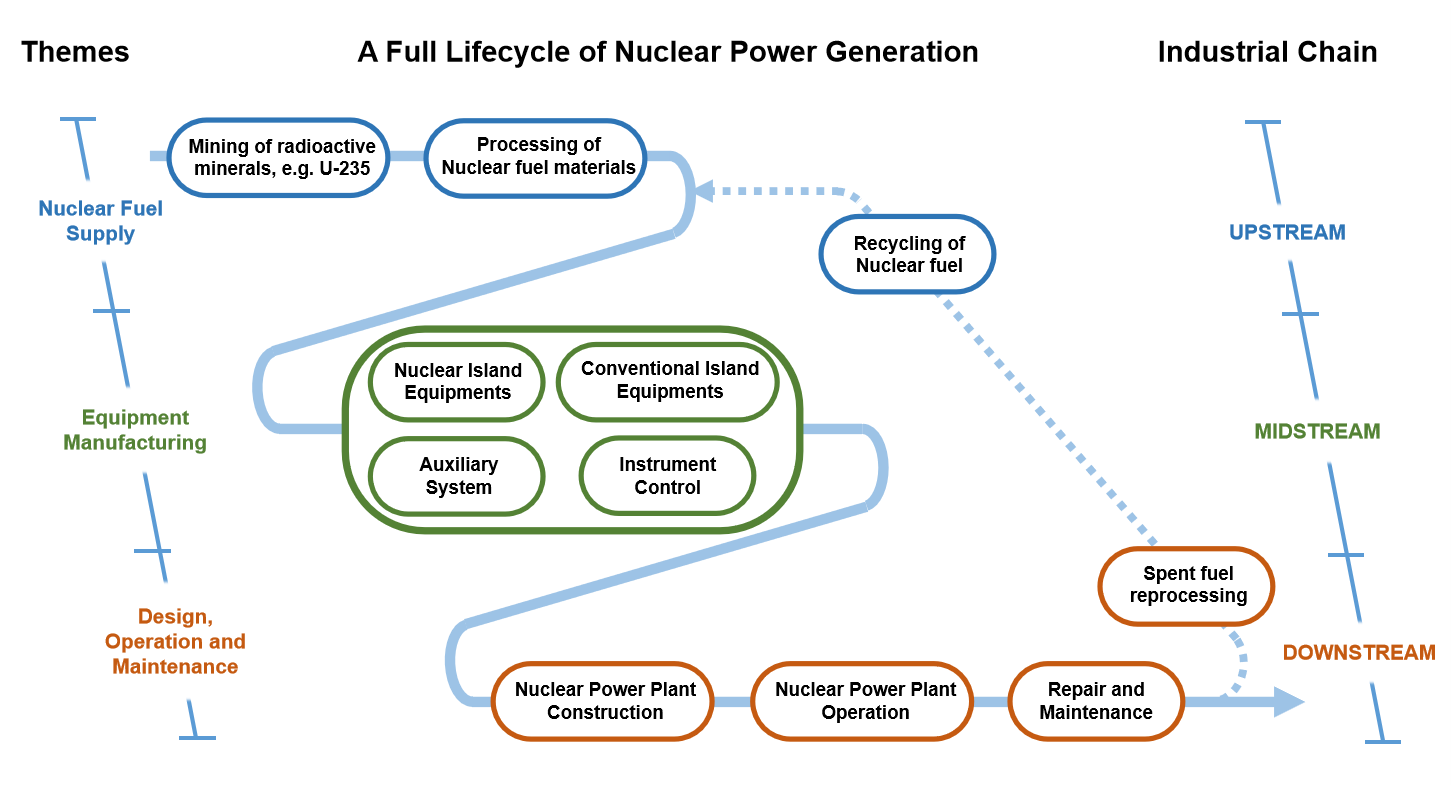}
\caption{A full life-cycle of NPG, with themes and works in each stage corresponding to the upstream, midstream, and downstream of the industrial chain.}
\label{consGraphs1}
\end{figure*}

\subsection{Full life-cycle intelligence of NPG}
\label{lifecycel}
NPG, as the most common and complete civilian nuclear industry, its full life-cycle involves multiple stages, such as the exploration and mining of radioactive minerals, the research, development, and manufacturing of nuclear fuel, and the processing of spent fuel and nuclear waste, each of which has its unique requirements for intelligentization \cite{tanaka2015size}. Applying AI to each stage of the nuclear power life-cycle and extending it to other nuclear energy utilization scenarios undoubtedly has tremendous development potential. Figure~\ref{consGraphs1} shows the various stages of the life-cycle of NPG, which also belong to the upstream, midstream, and downstream of the nuclear power industrial chain. 

Early work in the nuclear power industry mainly involves the exploration and mining of naturally occurring radioactive minerals such as uranium-235, and the processing of nuclear fuel to manufacture core components such as fuel rods. If the reprocessing of spent fuel is also considered, the recycling of nuclear fuel is also part of the upstream of the industrial chain. The mid-term work mainly involves the manufacturing of core components of nuclear power plants, including three types of nuclear power equipment: nuclear island equipment, conventional island equipment, and auxiliary systems, as well as corresponding instrument control. The later stage, which is downstream of the nuclear power industry chain, mainly involves meeting the high safety requirements of nuclear power plant construction, operation, maintenance, and the reprocessing of spent nuclear fuel \cite{tang2022deep}. After an in-depth analysis of the above-mentioned full life-cycle of NPG, this study first summarizes some application scenarios that can introduce AI. Moreover, it pays special attention to scenes in which DL methods for finite samples, e.g., SSL, FSL, ZSL, and OSR, may be required as follows.

\begin{figure*}[t]%[htbp]
\centering
\includegraphics[width=0.85\textwidth]{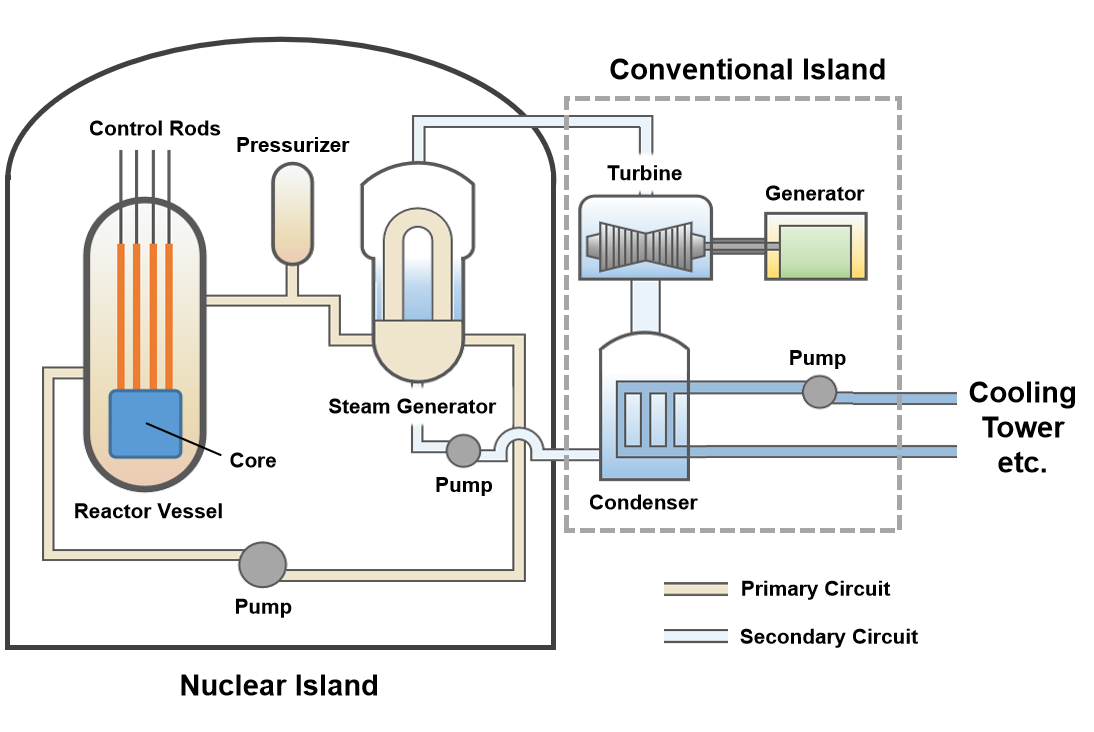}
\caption{Illustration of a nuclear power plant with Pressurized Water Reactor (PWR). The full system consists of the equipment and the auxiliary system in the nuclear and conventional islands, especially the two important cooling circuits. Meanwhile, instrument control is also a necessary part to ensure the normal operation of the power station.}
\label{consGraphs2}
\end{figure*}

\begin{itemize}
\item \textbf{Nuclear fuel supply.} AI techniques are introduced upstream of the nuclear power industrial chain, which mainly involves data analysis and procedure optimization for exploring and mining radioactive minerals, and processing and preparing nuclear fuel \cite{ali2020artificial}. For instance, DL can construct relevant knowledge graphs or enhance geophysical inversion to improve exploration efficiency, reduce working time, and mitigate risks and hazards during mining \cite{li2022overview}. Additionally, AI can manage massive data generated in nuclear fuel processing for real-time monitoring and adjustment of procedures \cite{wachter1987use}. Multi-objective optimization models can perfect processing operation parameters, leading to improved nuclear fuel rod yield, quality, and production efficiency \cite{andersen2023moogle}.

\item \textbf{Equipment manufacturing.} As shown in Figure~\ref{consGraphs2}, massive structured, unstructured, and semi-structured data are generated in the design, production, and operation of the nuclear island, conventional island, auxiliary systems, and instrument control equipment in the midstream \cite{tang2022deep}. DL can be used to analyze and process these data to establish efficient analysis and decision-making systems. In addition, considering the prominent structural issues in nuclear equipment manufacturing, i.e. high labor costs with low productivity, as well as the difficulty in controlling the operating and production environment of equipment, AI can be introduced for intelligent detection and recognition, to replace repetitive manual operations \cite{song2022online, song2023comparative}.

\item \textbf{Design, operation, and maintenance.} For the downstream works, including the construction, operation, and maintenance of nuclear power plants, establishing an integrated and intelligent platform for designing and constructing power plants can utilize the rising Physics-Informed Machine Learning (PIML) to support Multiphysics simulation and virtual design of the reactor and the whole system \cite{karniadakis2021physics}. To conduct regular inspections, environmental monitoring, and emergency response in hard-to-reach areas during power plant operation, maintenance, and repair stages, intelligent devices embedded with AI can be used \cite{huang2023review}. The application of DL methods for finite samples is crucial for timely anomaly detection or fault prediction with as few abnormal/faulty samples as possible. 
\end{itemize}  
%Furthermore, DL can model the highly nonlinear relationship between operating parameters and uranium utilization efficiency during the reprocessing of spent fuel, allowing for real-time optimization and increased recycling efficiency.

In general, many aspects throughout the full life-cycle of nuclear power can introduce AI to improve their intelligence. However, it should be noted that, like many other industrial scenarios aiming to introduce AI approaches, although the operation of nuclear power plants generates a large amount of data quickly and in real-time, the industrial systems do not wish to malfunction. As a result, a substantial portion of industrial data actually have characteristics of long-tailed class distribution \cite{zhang2023deep}, sample imbalance \cite{zhan2023hybrid}, and domain shift \cite{li2020deep}. In the subsequent sections, the data characteristics in NPG will be analyzed, and the theories and effectiveness of finite sample learning in solving these issues will be presented and discussed later.

\begin{figure}[t]%[htbp]
\centering
\includegraphics[width=0.98\textwidth]{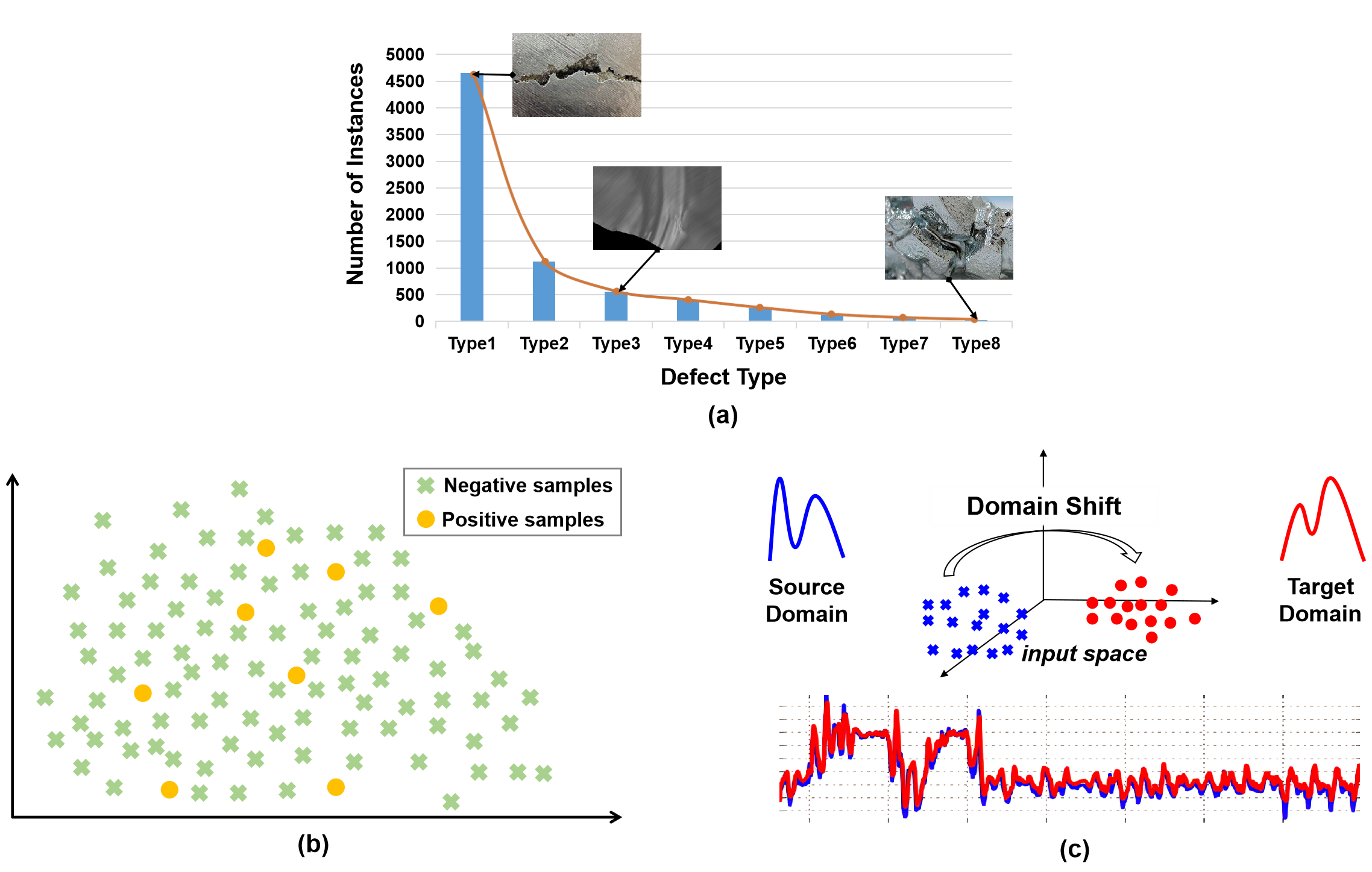}
\caption{The illustration of (a) long-tailed distribution, (b) sample imbalance, and (c) domain shift in NPG.}
\label{fig:data_charac}
\end{figure}

% \begin{figure}[t]
%   \centering
%   \subfloat[Long-tailed class distribution\label{fig:longtail}]{
%     \includegraphics[width=0.32\textwidth]{img/longtail.png}
%   } 
%   \subfloat[Sample imbalance\label{fig:imbalance}]
%   {
%     \includegraphics[width=0.32\textwidth]{img/imbalance.png}
%   }  
%   \subfloat[Domain shift\label{fig:domain}]{
%     \includegraphics[width=0.32\textwidth]{img/domain shift.png}
%   }
%   \caption{The illustration of long-tailed distribution, sample imbalance, and domain shift in NPG.}
%   \label{fig:data_charac}
% \end{figure}

\section{Characteristics of Data in NPG}
\label{data}
With the rapid development of big data technology and multi-source sensing, the scale and structure of distributed Industrial Internet of Things (IIoT)  are becoming increasingly complex. The massive amount of data brought by IIoT is characterized by multiple sources, heterogeneity, high dimensionality, and sparsity \cite{yu2022heterogeneous}. Generally, there are two main types of data in IIoT: visual data for monitoring operational environments and equipment conditions, and sequence data such as literature and sensor signals. Traditional heterogeneous data fusion methods and multi-source data processing methods based on Machine Learning (ML) often fail to address the characteristics of subjective elements, data redundancy, and fuzzy multi-source signal fusion strategies brought by these data characteristics, which leads to weak anti-interference and generalization abilities \cite{zhang2022bearing}. Recently, Deep Learning (DL) methods have achieved impressive results in some related classic tasks, especially in Computer Vision (CV) and Natural Language Processing (NLP) \cite{lecun2015deep}. However, when applying these high-performing DL models to specific NPG applications, such as defect detection, fault diagnosis, and risk warning, their performance often degrades due to the unique data characteristics of NPG data. As shown in Figure \ref{fig:data_charac}, in addition to multi-source heterogeneous and sparse diversity, there are also several special data characteristics in NPG, including long-tailed class distribution \cite{zhang2023deep}, sample imbalance \cite{zhan2023hybrid}, and domain shift \cite{li2020deep}. The subsequent analysis will focus on these significant data characteristics of NPG.

\begin{itemize}
\item \textbf{Long-tailed class distribution.} In practical scenarios, training samples often exhibit an imbalanced class distribution, where a small number of classes have a large number of samples while the rest have only a few. This results in a long-tailed class distribution where there are a few instances occurring frequently and a vast number of instances occurring infrequently \cite{zhang2023deep}. This class imbalance is a common issue in industrial intelligence applications in NPG. As shown in Figure \ref{fig:data_charac}(a), in the task of surface defect detection in metal equipment, 'Rust' is the dominant class, while 'Broken', 'Folds', 'Cracks', and others are the tail classes \cite{he2021real}. The long-tailed class distribution is also common in fault diagnosis, a classic industrial intelligence task in NPG. Machines often operate under normal conditions, and the probability of various faults is different, leading to extreme class imbalance and long-tailed distribution between different health conditions \cite{chen2022imbalance}. The long-tailed class distribution often limits the practicality of DL-based methods, as they tend to be biased towards dominant classes and perform poorly on tailed classes.

\item \textbf{Sample imbalance.} Unlike long-tailed distributions, sample imbalance often refers to situations where the distribution of the target variable in the dataset is heavily skewed towards one class. In binary classification tasks, if one class has significantly more samples than another class, the dataset is considered unbalanced. Data-driven abnormal working condition warning is an important component of industrial process monitoring in NPG, which distinguishes abnormal conditions from normal process data by using easy-to-measure process variables \cite{fan2021imbalanced}. As shown in Figure \ref{fig:data_charac}(b), sample imbalance is reflected in the fact that there are far more samples under normal operating conditions (negative samples) than under abnormal operating conditions (positive samples). Similar to the long-tailed class distribution, sample imbalance can also affect the generalization and robustness of the model, as the model may become biased towards the majority class and perform poorly on the minority class.

\item \textbf{Domain shift.} Recently, many data-driven approaches based on DL have been successfully developed for basic industrial intelligence tasks. However, these methods rely on the assumption of independent and identically distributed (i.i.d.) data, i.e., the distribution of training data from the source domain is the same as that of testing data from the target domain, which is often not satisfied in practical industrial applications \cite{han2020deep}. Factors like environmental noise and changing operational conditions can cause a domain shift between the training and testing data. For example, data collected from a simulation experiment or one piece of equipment may not be representative of the test data from another piece of equipment or the actual scenarios. Even data collected from different locations on the same device can exhibit distribution discrepancies. Additionally, most data-driven methods assume a closed-set assumption where the training and testing data belong to the same class set \cite{chen2022open}. However, in industrial scenarios, the operating environment and conditions may change over time, leading to the generation of data with new characteristics and distributions that belong to unknown classes, known as the open-set domain adaptation problem \cite{panareda2017open}. As shown in Figure \ref{fig:data_charac}(c), the domain shift between the training data (source domain) and testing data (target domain) presents unique challenges in practical applications \cite{muller2022unsupervised}.
\end{itemize}  

\section{Deep Learning with Finite Samples}
\label{deeplearning}
The unique data characteristics, i.e., long-tailed class distribution, sample imbalance, and domain shift, pose significant challenges for DL-based methods and require specialized techniques such as continual learning \cite{li2023large}, domain adaptation \cite{oyewole2022controllable}, transfer learning \cite{fan2020statistical}, multimodal learning \cite{liu2023transformer}, and generative modeling \cite{dumas2022deep} to address them. In this paper, based on our previous research work, we focus on reviewing the research progress of deep learning from the finite sample perspective, including SSL, FSL, ZSL, and OSR. 

\subsection{Small-Sample Learning}
\label{sec:smallsample}
In industrial practice, especially in the nuclear power industry, where high equipment safety and reliability are required, factories operate mostly under normal conditions and have limited samples of faults and defects. Additionally, annotating data professionally and extensively can also be challenging. The three characteristics of NPG data mentioned earlier, i.e., long-tailed class distribution, sample imbalance, and domain shift, are essentially challenges that arise due to the small sample \cite{chen2022imbalance}. Specifically, in the long-tailed class distribution, the samples of the tail classes are too small, while in sample imbalance, the samples of minority classes are too small. Domain shift often occurs due to incomplete data annotation and limited training samples. When the sample size decreases, data-driven DL methods often encounter problems such as under-fitting or over-fitting, leading to a decrease in model performance. Thus, small-sample learning has emerged as a hot research topic in NPG \cite{qian2022development}.

Small-sample learning refers to the task of training a neural network on a dataset with a small number of samples, whether labeled or not. For a given neural network $\mathcal{N}$, traditional deep learning algorithms require a dataset with $\mathcal{D}$ samples to train the network to convergence. The goal of small-sample learning is to design an algorithm $\mathcal{A}$ that can make neural network $\mathcal{N}$ achieve or approach convergence on a dataset with $\mathcal{S}$ samples, where $\mathcal{S} \ll \mathcal{D}$ \cite{liu2019network}. The limited amount of data makes it challenging to train a deep neural network. Currently, there are three main research directions to improve the performance of neural networks in small-sample learning tasks: knowledge transfer, data augmentation, and algorithmic improvement. Knowledge transfer involves borrowing experiences from pre-trained models through transfer learning and multi-task learning to solve new problems \cite{zhang2018fine}. Data augmentation involves obtaining more training data through specific methods to augment the training dataset. Data augmentation can enhance the network's robustness by introducing different types of data invariance, such as using translation, flipping, shearing, scaling, reflection, cropping, and rotation on images. Additionally, in small-sample learning, using generative models to directly generate augmented data is also a hot topic of research \cite{antoniou2018augmenting}. Algorithmic improvement involves using learning techniques such as dropout to improve the efficiency of neural network training, or using other learning methods such as incremental learning to give the network memory capabilities \cite{Doyen2018Online}.

\begin{figure*}[t]%[htbp]
\centering
\includegraphics[width=0.93\textwidth]{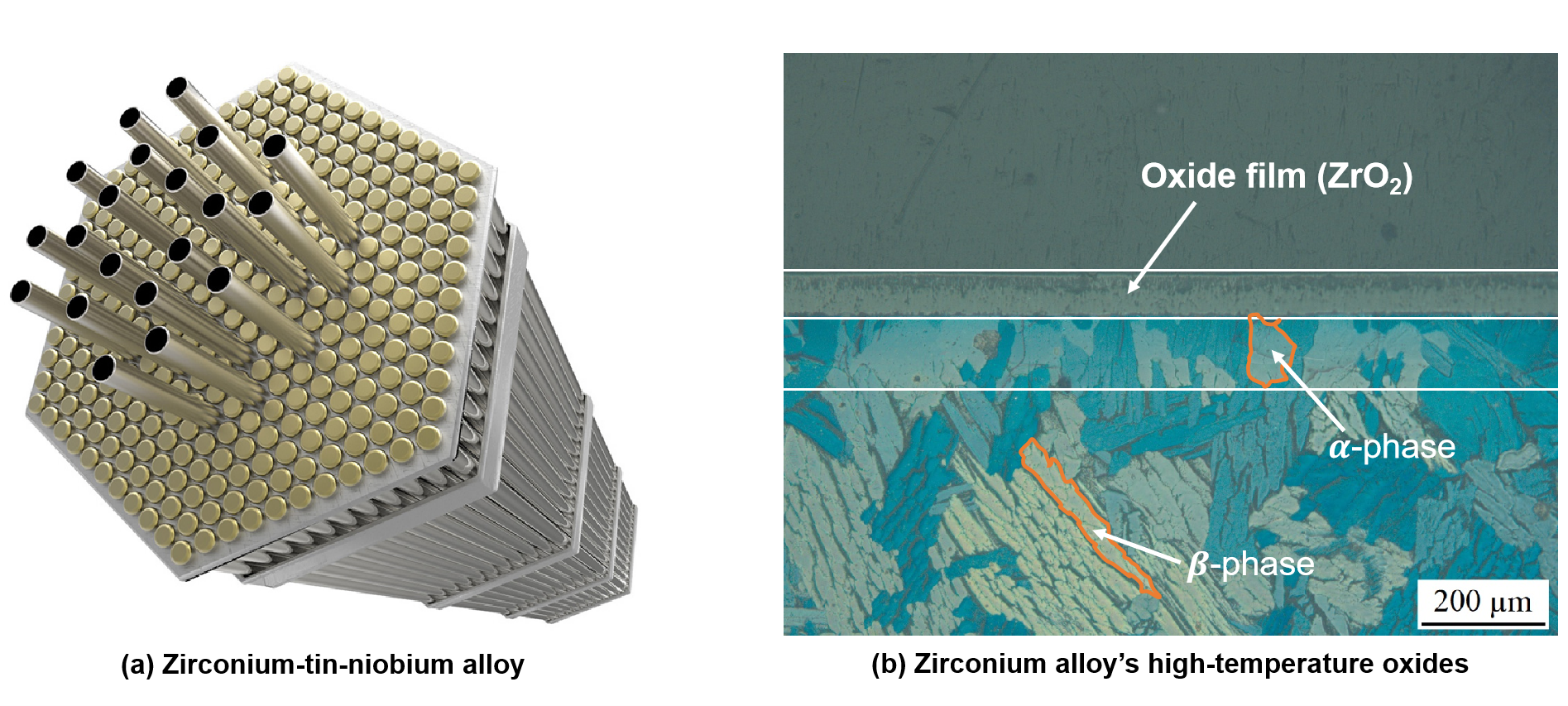}
\caption{Application of the zirconium alloy in nuclear power plant as the cladding tube material and microstructure of the zirconium alloy’s high-temperature oxides observed with scanning electron microscopy (model: ZEISS SUPRA55).}
\label{consGraphs7}
\end{figure*}

% \begin{figure}[t]
%   \centering
%   \subfloat[Zirconium-tin-niobium alloy\label{fig:zir1}]{
%     \includegraphics[width=0.4\textwidth]{img/zir1.png}
%   } 
%   \subfloat[Zirconium alloy’s high-temperature oxides\label{fig:zir2}]
%   {
%     \includegraphics[width=0.4\textwidth]{img/zir2.png}
%   }  
%   \caption{Application of the zirconium alloy in nuclear power plant as the cladding tube material and microstructure of the zirconium alloy’s high-temperature oxides observed with scanning electron microscopy (model: ZEISS SUPRA55).}
%   \label{consGraphs7}
% \end{figure}

%(a) Application of the zirconium alloy in nuclear power plant as the cladding tube material and (b) microstructure of the zirconium alloy’s high-temperature oxides observed with scanning electron microscopy (model: ZEISS SUPRA55).

\subsection{Any-Shot Learning}
\label{sec:zershot}
Despite the numerous methods proposed to solve the small-sample problem, there are still two challenges in NPG that make the performance of these small-sample learning methods perform poorly. One challenge is the zero-shot problem, where DL models cannot recognize instances of new classes that did not appear during the training procedure. The other challenge is the few-shot problem, where labeled samples are extremely scarce (less than 10 per class) \cite{zhuo2021auxiliary}. Most DL methods perform poorly on any-shot problems, i.e., zero-shot and few-shot problems, due to the limited intra-class information. The construction and training process of these DL models is only based on the dataset itself, which limits the introduction of external auxiliary information outside the training set. For instance, when encountering surface defects that are not in the training set, these DL methods cannot recognize these unseen defects. When it comes to the few-shot problem, the situation may be better because there are no unseen defects, and some semi-supervised methods can fully utilize large unlabeled samples. However, in cases where samples are extremely scarce, most of them still perform poorly because the information provided by a few samples is insufficient for these methods to learn accurate data distributions. Recently, with the development of Zero-Shot Learning (ZSL) and Few-Shot Learning (FSL), any-shot learning has gradually been used to solve some extremely challenging tasks in the intelligent maintenance of NPG \cite{feng2020fault}.

ZSL aims to recognize instances of unseen classes that have no available labeled data for training. The classes in the training set (seen class $\mathcal{S}$) and those in the testing set (unseen class $\mathcal{U}$) are disjoint, i.e., $\mathcal{Y}_\mathcal{S} \cap \mathcal{Y}_\mathcal{U} = \emptyset$ \cite{ali2009Describing}. To classify unseen classes, ZSL assumes that there is a common semantic space where the visual information and class labels of both the seen and unseen classes can be projected. Then, the semantic representation can be leveraged to construct the visual-semantic embedding space which transfers knowledge from seen to unseen classes \cite{AWA2-gbu}. Many effective methods have been proposed to build the visual-semantic embedding space, with cross-modal embedding models \cite{TANG2020105490} and generative models \cite{tang2021zero} being the two mainstream methods. The cross-modal embedding models often project visual space to the semantic space or vice versa or project both features to a common embedding space. Generative models, such as Generative Adversarial Nets (GANs) \cite{Ian2014Generative}, have been recently developed, leading to more studies focusing on ZSL methods based on generative models. With the generated visual features, ZSL can be improved by reformulating it as a fully supervised classification task.

FSL learns classifiers $f: x_i \to y_i$ given only a limited number of labeled training examples of each class. FSL is typically framed as an N-way-K-shot classification problem, with the training set $D_{train} = {(x_i,y_i)}_{i=1}^I$ consisting of $I = KN$ examples from $N$ classes, each having $K$ examples \cite{wang2020generalizing}. While FSL shares similarities with semi-supervised learning and imbalanced classification problems, it stands apart due to the significantly lower quantity of labeled instances, making the task notably more challenging. The fundamental technique of FSL is to leverage the knowledge gained from a set of base classes with abundant labeled data, and then generalize this knowledge to new tasks that contain only a few samples with supervised information \cite{tang2023privacy}. The main challenge in FSL is the unreliability of the empirical risk minimizer. Various FSL methods have been devised to address this, using prior knowledge from three perspectives: 1) data: prior knowledge is used to augment the supervised experience; 2) model: prior knowledge is used to reduce the size of the hypothesis space; 3) algorithm: prior knowledge is used to modify the search for the optimal hypothesis within the given hypothesis space. Meta-learning is a crucial technique in tackling FSL problems, as it enables the model to learn how to solve tasks during training, and it can then generalize to new tasks that it has not encountered before \cite{sun2019meta}. Another strategy employed in FSL is based on metric learning. This approach focuses on learning an embedding function, using the similarity metrics of embeddings to classify instances \cite{jiang2020multi}.

\begin{figure}[b]%[htbp]
\centering
\includegraphics[width=0.85\textwidth]{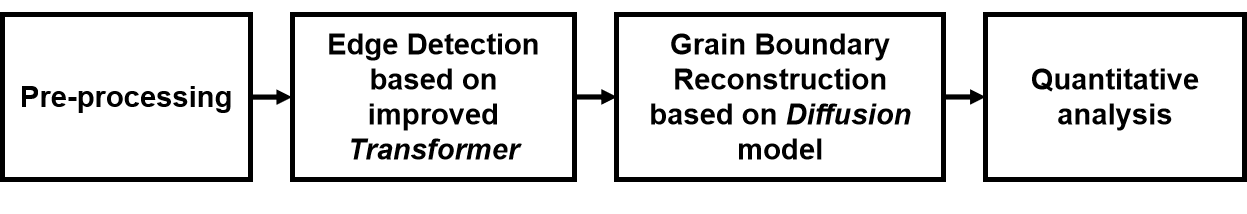}
\caption{The flowchart of a DL-based approach for solving the small sample problem in the automatic recognition task of zirconium alloy metallography.}
\label{consGraphs8}
\end{figure}

% \begin{figure}[t]%[htbp]
% \centering
% \includegraphics[width=0.8\textwidth]{img/flowofcase1.png}
% \caption{The flowchart of a DL-based approach for solving the small sample problem in the automatic recognition task of zirconium alloy metallography.}
% \label{consGraphs8}
% \end{figure}

\subsection{Open-Set Recognition}
\label{sec:openset}
In many real-world recognition tasks, it is often challenging to collect a comprehensive set of training samples that encompasses all possible classes. As a result, the training of a recognizer is typically limited by various objective factors. A more practical scenario is OSR, where the training dataset has incomplete knowledge of the world, and unknown classes can be encountered by the algorithm during testing \cite{muller2022unsupervised}. This requires the recognizer to accurately classify the seen classes, and effectively handle the unseen ones. In other words, classifiers need to be capable of recognizing known classes and detecting unknown ones. Let $V$, $\mathcal{R}_\mathcal{O}$, and $\mathcal{R}_\epsilon$ denote the training data, open space risk, and empirical risk, respectively. The main objective of OSR is to find a measurable recognition function $f \in \mathcal{H}$ \cite{scheirer2012toward}. Here, $f(x) > 0$ implies correct recognition and $f$ is defined by minimizing the \textit{Open Set Risk}, which is given by:
\begin{equation}\label{osr}
\begin{gathered}
\arg \min_{f \in \mathcal{H}} \{\mathcal{R}_\mathcal{O}(f) + \lambda \mathcal{R}_\epsilon(f(V))\},
\end{gathered}
\end{equation}
where $\lambda$ is a regularization constant. Compared to ZSL and FSL, OSR faces a more serious challenge. This is because OSR only has access to classes with distinctly labeled positive training samples, without any additional side-information such as attributes or a limited number of samples from classes without any information during training. Therefore, OSR requires more sophisticated techniques to address the challenge of recognizing unknown classes in real-world scenarios \cite{geng2020recent}. Researchers have explored the modeling of OSR from both discriminative and generative perspectives, with certain constraints \cite{perera2020generative}. Based on their modeling forms, these models can be further categorized into four main categories: Traditional Machine Learning (TML)-based and Deep Neural Network (DNN)-based methods from the discriminative model perspective, and Instance-based and Non-Instance-based Generation methods from the generative model perspective. 

\section{Case Studies}
\label{case}
As previously discussed, DL methods with finite samples hold immense potential for application in the full life-cycle intelligence of NPG. This is derived from their technical superiority and is further bolstered by the intrinsic attributes of industrial scenarios. In this section, we aim to provide a comprehensive understanding of completed works by presenting two case studies. The first case study involves the automatic recognition of zirconium alloy metallography using SSL based on the improved transformer \cite{huang2020improving} and diffusion model \cite{bansal2023universal}. Another case study focuses on OSR for signal diagnosis of machinery sensors in the industrial environment.

\begin{figure}[b]%[htbp]
\centering
\includegraphics[width=0.92\textwidth]{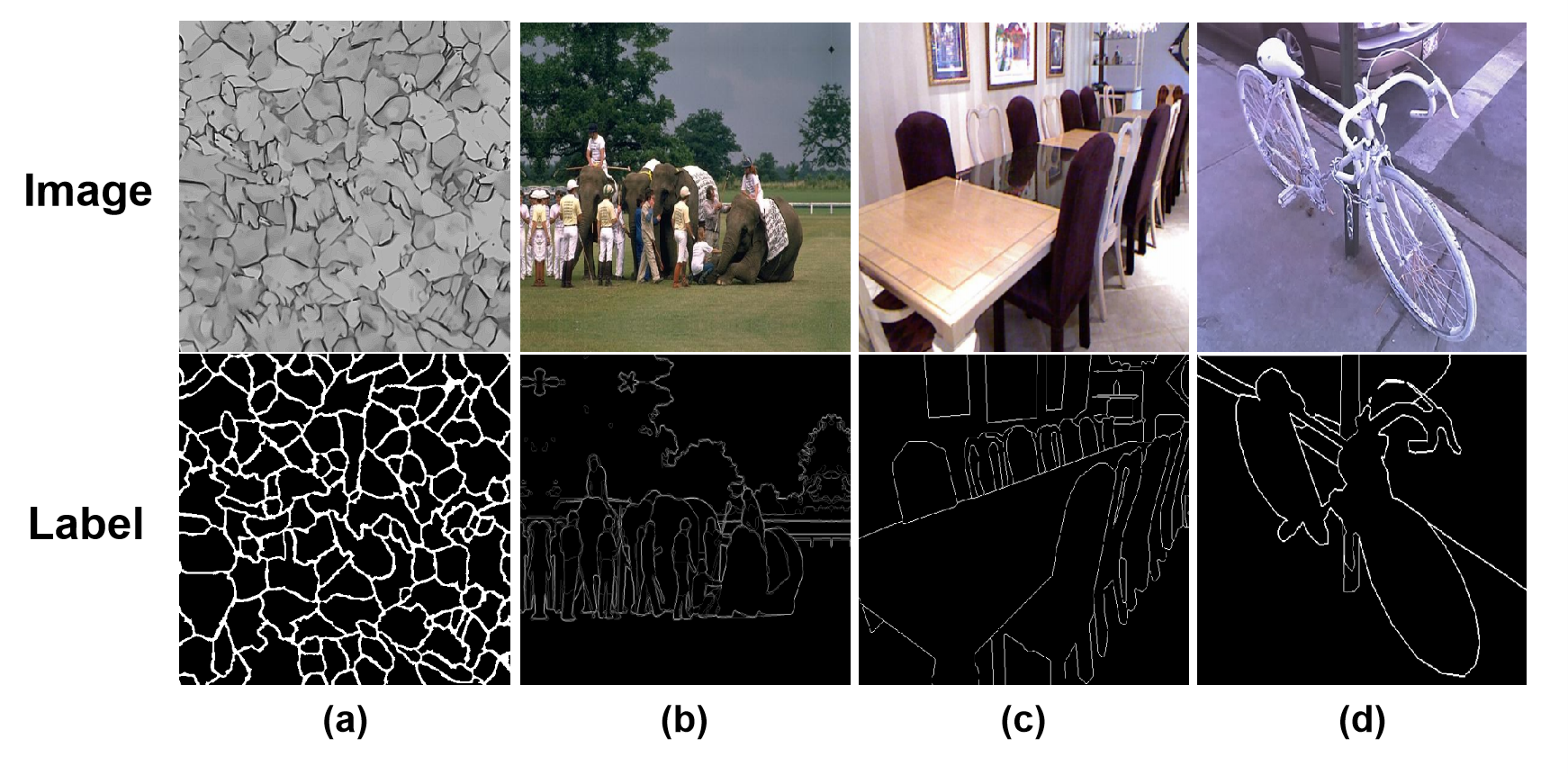}
\caption{The data pairs consist of images (up) and labels (down, object edges) in different datasets, i.e., zirconium alloy metallographic, BSBS500, NYUDv2, and Pascal-Context datasets.}
\label{consGraphs9}
\end{figure}

\subsection{Automatic Recognition of Zirconium Alloy Metallography}
In the nuclear industry, zirconium (Zr) alloys possess remarkable properties essential for constructing nuclear reactors, including high strength, excellent ductility, corrosion and oxidation resistance, and extremely low thermal neutron absorption cross-section \cite{sabol2000zirconium}. For this reason, they are commonly used to manufacture crucial components such as cladding materials (zirconium-tin-niobium alloy, see Figure~\ref{consGraphs7}(a)), deceleration materials (zirconium hydride), nuclear fuel (zirconium fluoride), and accessory materials (uranium alloy additives), etc. However, the Fukushima Daiichi nuclear power plant accident in 2011 exposed the potential for rapid oxidation failure of the zirconium alloy cladding material in a high-temperature steam environment caused by a Loss Of Coolant Accident (LOCA) \cite{syrtanov2022high}. As a result, intensive researches on the high-temperature oxidation of zirconium alloys and the development of targeted criteria for analysis and evaluation are crucial to mitigate such flaws and enhance nuclear power plant safety.

Figure~\ref{consGraphs7}(b) shows a typical microstructure of the high-temperature zirconia alloy observed by scanning electron microscopy, which is also referred to as a metallographic image of the metal alloy. Several different layers visible in the figure start at the top with the tabletop of the bench, followed by a thin oxide film consisting mainly of zirconium dioxide (ZrO$_2$). Beneath it lies the $\alpha$ phase of zirconium alloy ($\alpha$-Zr), also known as the low-temperature phase, which is more stable at lower temperatures. Meanwhile, the $\alpha$ phase exhibits stronger creep resistance and higher strength than the next $\beta$ phase ($\beta$-Zr), also known as the high-temperature phase, which is more stable at higher temperatures and has superior corrosion resistance \cite{wang2023low}. As the oxidation progresses, oxygen atoms cross the oxide film and react with the $\beta$ phase, which possesses a larger aspect ratio compared to the $\alpha$ phase. This procedure will gradually increase in the proportion of the $\alpha$ phase, with the oxide film becoming thicker. 

\begin{figure*}[t]%[htbp]
\centering
\includegraphics[width=0.985\textwidth]{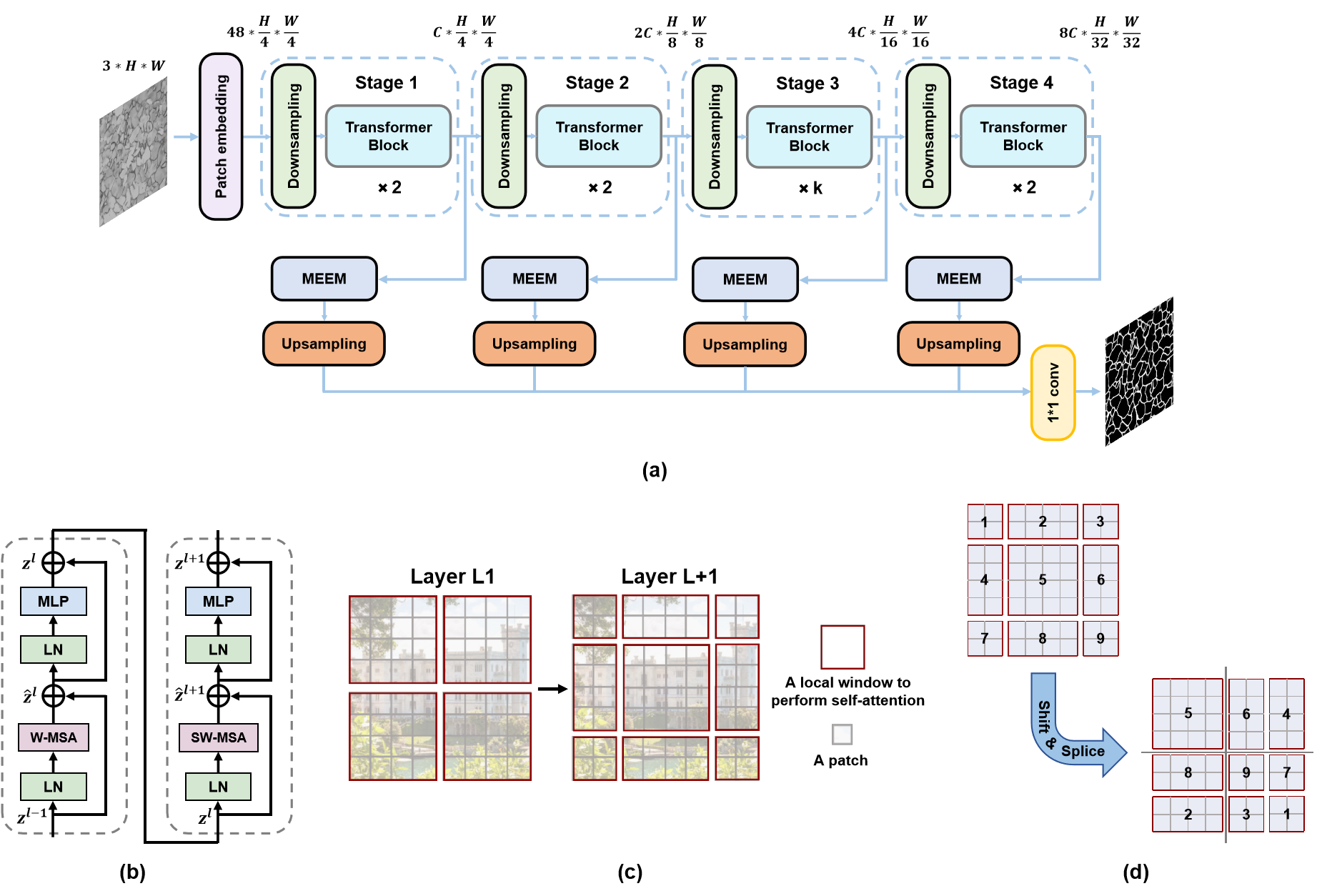}
\caption{(a) Illustration of the end-to-end edge detection network based on improved Transformer. To enhance the ability of extracting local features, a hierarchical structure that combines the strengths of both convolutional neural network (CNN) and Transformer is adopted: four downsampling layers are added to generate multi-level features that capture local information in the shallow layers and high-level semantic features in the deeper layers; (b) another design is the use of W-MSA in the previous Transformer block and SW-MSA in the next; (c) this allows for faster information flow and enables the model to capture semantic relationships between windows, e.g. the window in the middle of layer L+1 can exchange information with the four windows in layer L; (d) to retain the computational efficiency, nine windows would first be reassembled into four to perform the computation, and then split into nine again.}
\label{consGraphs10}
\end{figure*}

To systematically describe and study the evolution of zirconium alloy, it is essential to accurately identify and analyze the thickness of the oxide film and the proportion of $\alpha$ and $\beta$ phases in the microstructure at various oxidation temperatures and times from similar metallographic images. Currently, identifying the metallographic morphology of zirconium alloys and determining the resulting structures, status, and properties are mostly achieved qualitatively through visual inspection and manual counting of the relevant indicators in the images. However, this practice is not only inaccurate but also time-consuming and laborious. Hence, it is significant to develop a fast and accurate intelligent recognition algorithm for the metallography of zirconium alloys, and the corresponding application program based on this should be further bred. In this case study, the DL approach with powerful feature extraction capability is designed to accomplish a series of computer vision tasks such as target recognition and image generation, thereby achieving efficient recognition of microstructures of high-temperature zirconium oxide alloys. Specifically, metallographic image recognition and quantitative analysis are carried out in four stages as elaborated below. Figure~\ref{consGraphs8} provides a certain overview of each stage, including pre-processing, edge detection, grain boundary reconstruction, and quantitative analysis. Thus, solving the outstanding small sample problem in metallographic recognition of zirconium alloy, and also providing feasible solutions to the difficulties such as incomplete grain boundary edges.

\subsubsection{Pre-processing of metallographic images}
The first stage is aimed at pre-processing the metallographic images of zirconium alloys. The main objective of zirconium alloy metallographic recognition is to obtain the areas and proportions of different phases, and this goal is very dependent on the accurate detection of grain boundaries. The quality of image annotation in DL very much affects the final performance of the network model. Considering that there is no publicly available zirconium alloy metallographic dataset, zirconium alloy metallographic images were first acquired by using a scanning electron microscope. Figure~\ref{consGraphs9} shows the data pairs consisting of images (up) and labels (down, object edges) in different datasets. In Figure~\ref{consGraphs9}(a), after the oxide zirconium alloy was embedded and processed into metallographic specimens, 50 metallographic images (in color) were obtained with a resolution of 1200×900 after grinding, polishing, and etching operations. Although the above metallographic images possess high resolution, the number of images acquired is insufficient to develop an accurate recognition model. Additionally, the size of the original images is too large and does not facilitate training and testing. Thus, the images were subjected to grayscale processing and uniformly sliced to a size of 512×512. To increase the size of the dataset, the images were augmented through rotation, flip, random cropping and scaling, saturation variation, and contrast variation. These transformations resulted in a total of 4200 labeled metallic image pairs, which were then divided into a training set, validation set, and test set at an 8:1:1 ratio for training the DL models. 

\begin{figure}[b]%[htbp]
\centering
\includegraphics[width=0.5\textwidth]{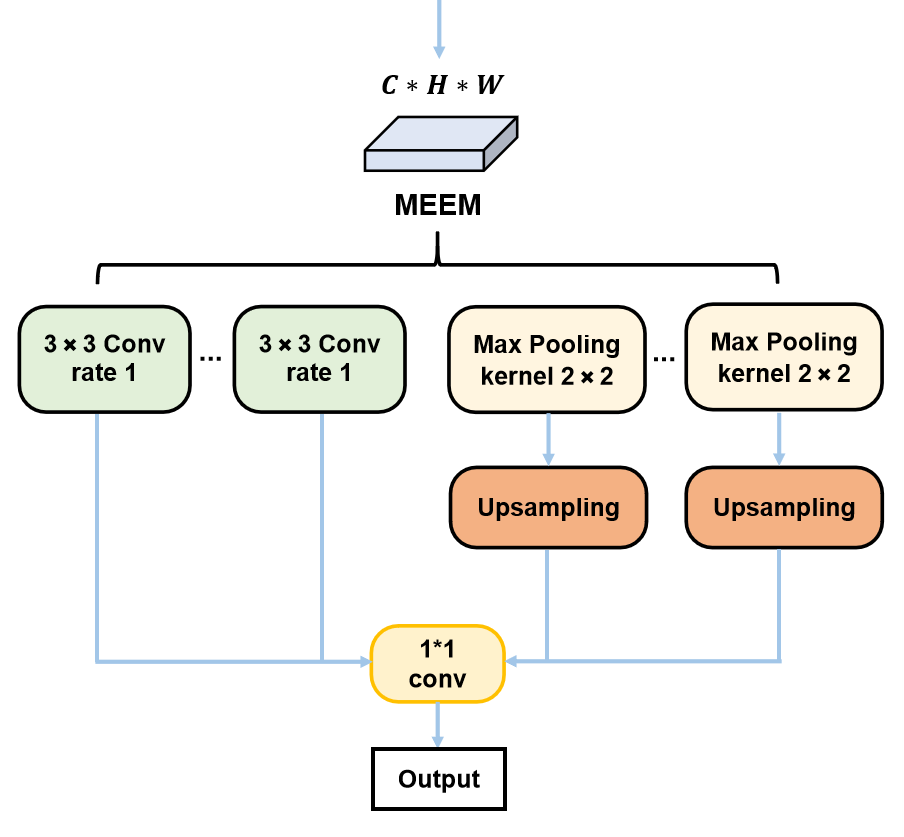}
\caption{The workflow of MEEM: the input of this module is the output features from each stage of the backbone network, which are further obtained by several dilated convolutions and max pooling to get features with larger receptive fields. The resulting features from the max pooling layers are up-sampled to match the size of the input features. Finally, the fused output features are obtained using a 1×1 convolution.}
\label{consGraphs11}
\end{figure}

In addition, the preparation and collection of metallographic images of zirconium alloys are challenging, and manual annotation is expensive, taking up to three hours for a single image and even longer if the grain size is small. Therefore, transfer Learning was introduced to enhance the neural network's performance. In the subsequent works, the network model was trained following the Richer Convolutional Features (RCF) strategy to converge on a dataset of edge detection comprising a blend of BSBS500, NYUDv2, and Pascal-Context datasets, and then fine-tuned on the prepared dataset of metallographic images of zirconium alloys \cite{gao2021end}. As shown in Figure~\ref{consGraphs9}, Figure~\ref{consGraphs9}(b) is the BSDS500 dataset, which consists of 500 images, with 200 for training, 100 for validation, and 200 for testing. This dataset is typically augmented for actual use. Figure~\ref{consGraphs9}(c) displays the NYUDv2 dataset, which consists of 1,440 images, with 381 for training, 414 for validation, and 654 for testing. Figure~\ref{consGraphs9}(d) depicts the Pascal-Context dataset, which consists of 10,103 images, with 4,998 for training and 5,105 for testing.

\begin{figure}[t]%[htbp]
\centering
\includegraphics[width=0.95\textwidth]{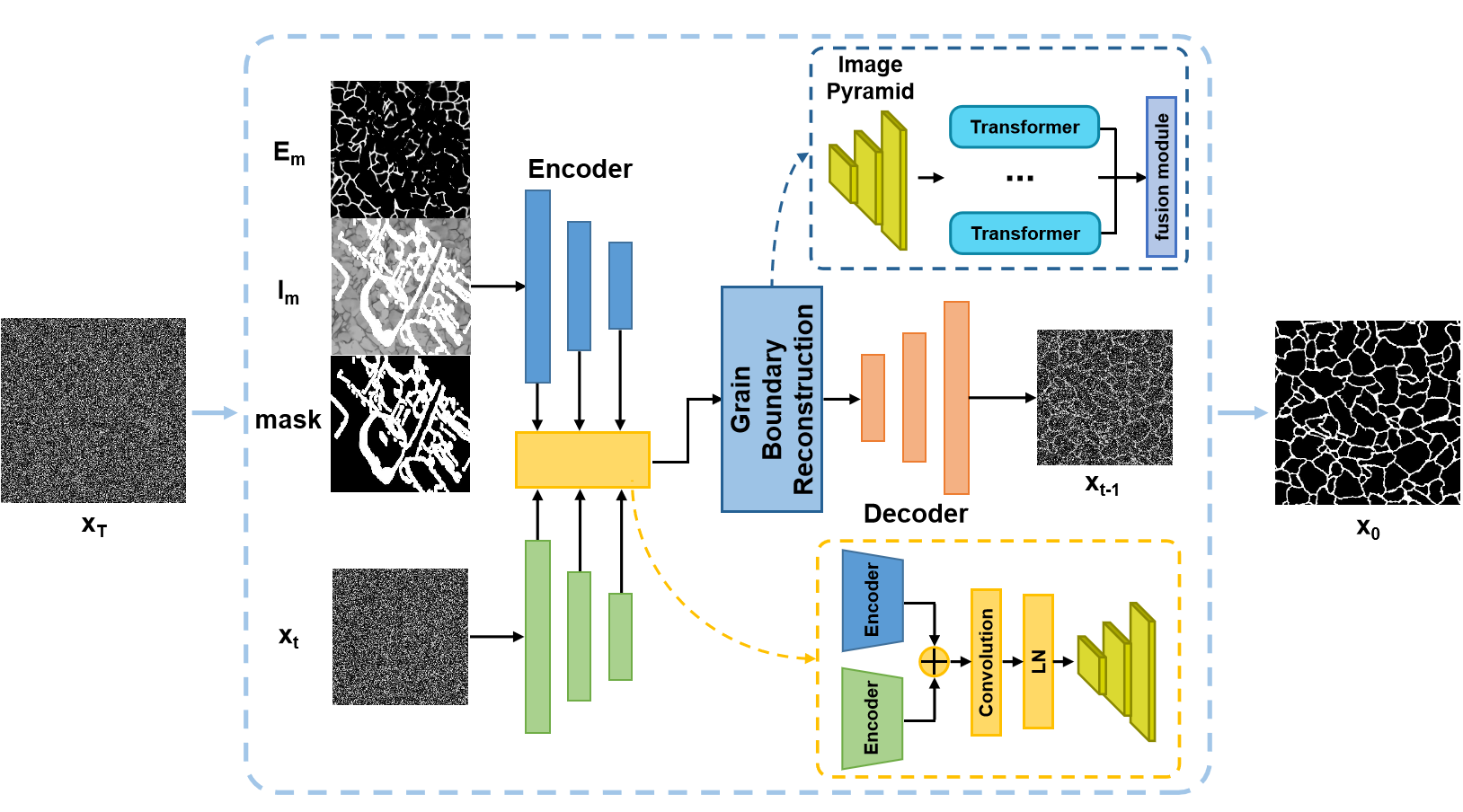}
\caption{The grain boundary reconstruction method based on diffusion model: the feature $X_{fuse}$ guided fusion by four semantics fusions from the encoder will be input into the Transformer block as an image pyramid to extract the overall structural features. Meanwhile, the multi-scale feature fusion module after the Transformer block accepts multiple feature maps with different sizes and then fuses them together as output.}
\label{consGraphs12}
\end{figure}

\subsubsection{Edge detection based on improved Transformer}
To address key issues in edge detection, particularly in metallographic images such as the small grain size in zirconium alloy metallography that requires more detailed inspection, an end-to-end edge detection network based on improved Transformer has been developed, which is called EDTR. Figure~\ref{consGraphs10}(a) depicts the overall structure of EDTR, which is divided into four stages, each utilizing multiple Transformer blocks as shown in Figure~\ref{consGraphs10}(b). It is worth noting that a crucial design of the new model is the use of window-based self-attention instead of global self-attention, resulting in a significant reduction in computational effort. In addition, a shifted window based multi-headed attention (SW-MSA, Figure~\ref{consGraphs10}(c)) is designed to work alongside normal window based multi-headed attention (W-MSA) to ensure that global information in the image is not overlooked. This approach facilitates faster information flow in the model and enables the capture of semantic relationships between windows. Figure~\ref{consGraphs10}(d) illustrates a fast computation method for offset window self-attention. In addition, a new module named Multi-scale Edge Extraction (MEEM) was also proposed to improve the Transformer's ability to extract small target features. The module is designed based on the dilated convolution technique: with different dilation rates, features of various sizes of receptive field can be adequately obtained using the dilated convolution. Figure~\ref{consGraphs11} demonstrates the MEEM procedure, which involves a series of integrated dilation convolution and max pooling to extract multi-scale edges.

Meanwhile, to address the class imbalance and difficulty in distinguishing pixels around edges in metallographic images, a new loss function with dynamic weight control was designed. While traditional edge detection uses a binary cross-entropy loss function, the new function is inspired by HED's use of coefficients to balance positive and negative samples. It sets the coefficients of indistinguishable samples as positive samples, increasing their weights as follows: 
\begin{equation}\label{HED}
\begin{gathered}
IBL(y,\hat{y})= -(w_{positive}*y*log(\hat{y}) + w_{negative}*(1-y)*log(1-\hat{y})).
\end{gathered}
\end{equation}

However, pixels close to edges still experience difficulty in differentiation, so the scaling factor $tanh(\hat{y})$ was introduced. $tanh(\hat{y})$'s smoother curve promotes stable training, while its more dramatic value changes around 0 better discriminate background pixels. The final loss function supervises both the output of each block and the fusion of all outputs as follows:
\begin{equation}\label{finalloss}
\begin{gathered}
L=\sum_{i=1}^N(\sum_{j=1}^Sw_{j}*IBDL(y_{i}^j,\hat{y}_{i}^j) + w_{fuse}*IBDL(y_{i}^{fuse},\hat{y}_{i}^{fuse})),
\end{gathered}
\end{equation}
where $IBDL(y,\hat{y})=-(w_{positive}*y*tanh(1-\hat{y})*log(\hat{y}) + w_{negative}*(1-y)*tanh(\hat{y})*log(1-\hat{y}))$.

\begin{figure}[b]%[htbp]
\centering
\includegraphics[width=0.98\textwidth]{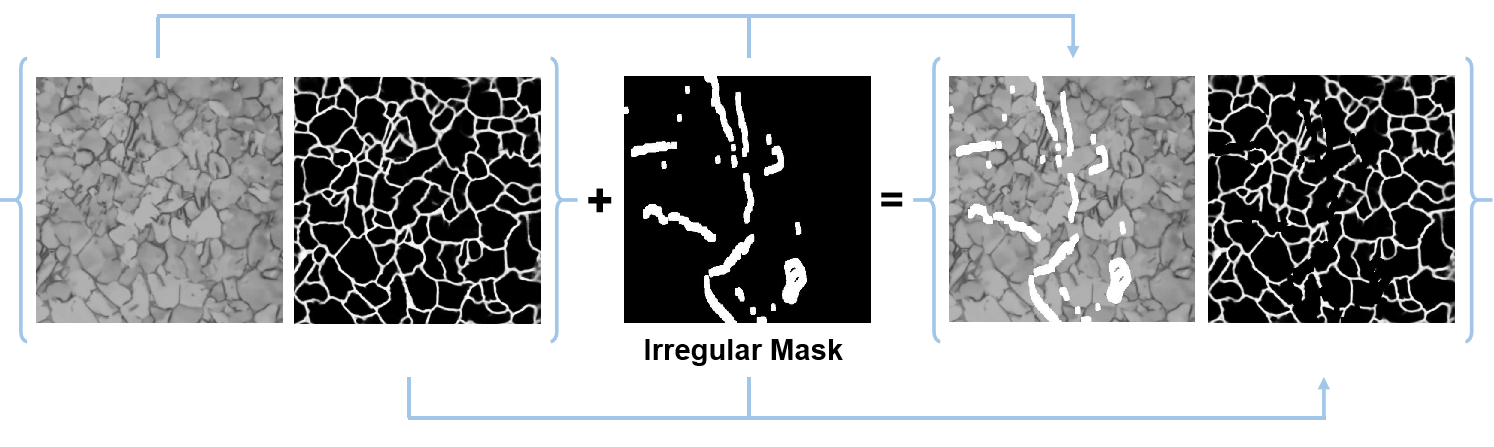}
\caption{Construction of the edge missing data.}
\label{consGraphs13}
\end{figure}

\subsubsection{Grain Boundary Reconstruction based on Diffusion model}
In the recognition of zirconium alloy metallographic images, another significant challenge arises from the defects that may be present in the images themselves. During the preparation of metallographic specimens, operational irregularities or inadequate cleanliness of the working environment can lead to unclear edges or even contaminated blocks in the images. Although the above-mentioned approach can effectively extract edges, it may still struggle to identify obscured areas. This is a problem faced by almost all current metallographic recognition algorithms. The traditional solution to this issue involves extending the edge along its direction until intersects. However, this approach contradicts the fact that metallographic images often contain numerous curved edges. To address this challenge, a grain boundary reconstruction method based on the diffusion model (Figure~\ref{consGraphs12}) was proposed, inspired by recent advancements in generative models, particularly Stable Diffusion, which has demonstrated significant advantages for this type of task.

The proposed method is named EdgeDiff, which leverages the concept of the diffusion model to define the reconstruction as a generative process. Specifically, a random noise image can be transformed into a reconstructed grain boundary with the aid of conditional information through several denoising steps. To further enhance the sensitivity of the model to the structural information of metallographic images, a grain boundary reconstruction module was incorporated. This module consists of an image pyramid and multiple Transformer blocks, with position encoding and axial Attention modifications employed to achieve a balance between performance and time efficiency. Additionally, a sampling acceleration algorithm was introduced to reduce the temporal overhead during the generation.

Specifically, for the preparation of the data set used for model training, inspired by the image inpainting, the irregular mask data set proposed by Liu et al. was adopted to construct more explicit edge missing data (Figure~\ref{consGraphs13})~\cite{10.1007/978-3-030-01252-6_6}. For the guidance by conditional information mentioned above, considering the high cost of training the original diffusion model based on random pictures and classifiers, the three types of conditional priors of the masks, the masked metallographic images, and the masked edge are used as input, features are extracted by the encoder and then fused with the features from the diffusion model encoder to obtain the image pyramid in the grain boundary reconstruction module. In particular, this module also enables the model to adaptively learn the mutual positional relationship between points in the feature through multiple position encodings, providing sufficient semantic information for grain boundary reconstruction. Finally, by adopting a non-Markov chain sampling algorithm, the speed of grain boundary reconstruction has been effectively improved.

\begin{table}[htbp]
\caption{Quantitative indicators for metallographic recognition.}
\label{quantitativeindicators} \scalebox{0.95}{
\begin{tabular}{ccc|ccc}
\hline\hline
\textbf{No.} & \textbf{Indicator}           & \textbf{Connotation}                                & \textbf{No.} & \textbf{Indicator}     & \textbf{Connotation}            \\
\hline
1            & Oxide film thickness         & Yes - output value/No                               & 2            & Area of $\alpha$ phase & pixels marked as $\alpha$ phase \\
3            & Proportion of $\alpha$ phase & $(area\ of\ \alpha) / (total\ pixels)$            & 4            & Area of $\beta$ phase  & pixels marked as $\beta$ phase  \\
5            & Proportion of $\beta$ phase  & $(area\ of\ \beta) / (total\ pixels)$             & 6            & Average grain size     & Ref. GB/T 6394-2017 {[}2{]}     \\
7            & Average circularity          & $(4 \Pi*grain\ area)/(perimeter^2)$ &              &                        &   \\ 
\hline\hline
\end{tabular}}
\end{table}

\begin{figure}[b]%[htbp]
\centering
\includegraphics[width=0.93\textwidth]{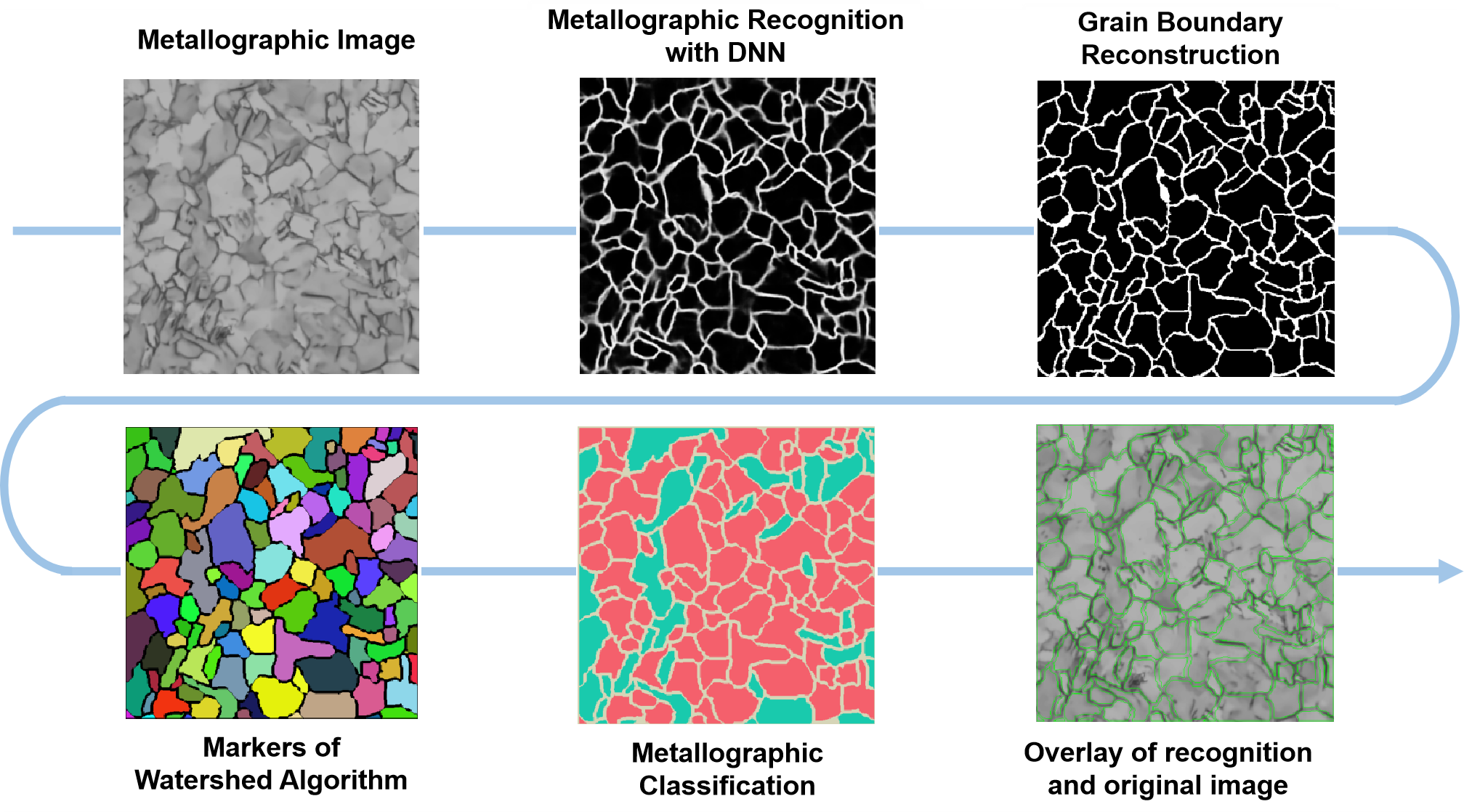}
\caption{A full procedure of the zirconium alloy metallographic recognition and the output of each step.}
\label{consGraphs14}
\end{figure}

\subsubsection{Quantitative analysis}
After completing the aforementioned tasks, the quantitative analysis of metallographic recognition can be conducted. Table~\ref{quantitativeindicators} presents the necessary quantitative indicators to calculate. In addition to the oxide film area, the area and proportion of the $\alpha$ phase, the area and proportion of the $\beta$ phase, the average grain size, and the average circularity are also computed. The average grain size reflects the average size of the grains, where a smaller average grain size typically indicates higher material strength and increased resistance to high-temperature oxidation. The average circularity serves as a supplementary indicator of the $\alpha$/$\beta$ ratio, reflecting the distribution of $\alpha$ and $\beta$ phases in the metallographic image. A circularity value closer to 1 indicates a higher presence of $\alpha$ phases, while a value closer to 0 suggests a higher presence of $\beta$ phases. To ensure the accuracy of quantitative analysis, the watershed algorithm is applied to segment incompletely closed edges. In the metallographic classification shown in Figure~\ref{consGraphs14}, the red phase represents the $\alpha$ phase with an area of 55,100.21 $\mu m^2$, the green phase represents the $\beta$ phase with an area of 12,819.38 $\mu m^2$, resulting in an $\alpha$/$\beta$ ratio of 4.30. The average grain size is graded as 9, and the average circularity is 0.57. Finally, a comparison is made between the edge mask and the original metallographic image to conclude the analysis.

\subsection{Open-Set Recognition for Signal Diagnosis}
\label{fewshotnuclear}
\begin{figure}[t]%[htbp]
\centering
\includegraphics[width=0.88\textwidth]{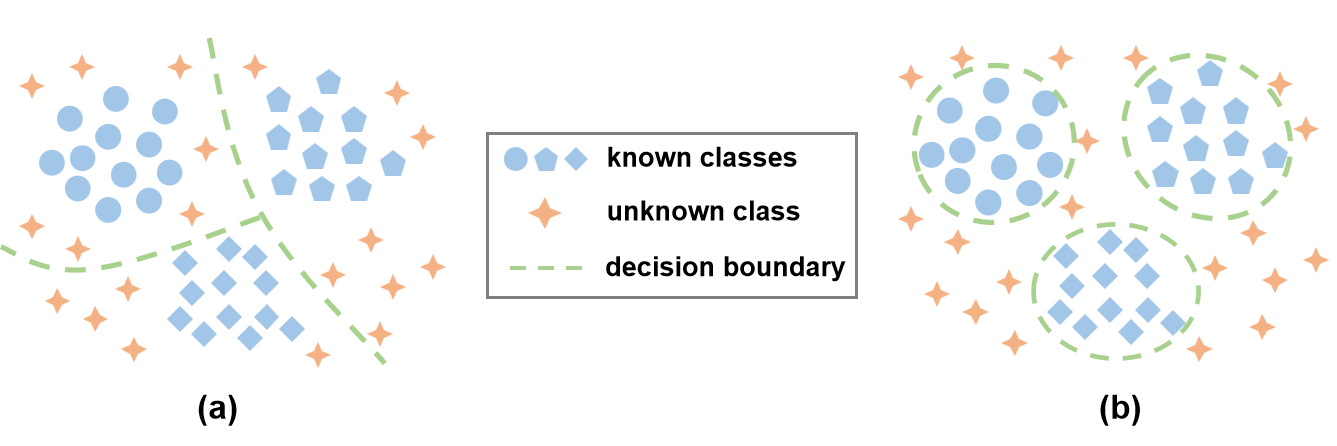}
\caption{Illustration of the (a) Closed-Set Recognition (CSR) and (b) Open-Set Recognition (OSR). In (a), the dashed line represents the decision boundary for each class. During testing, samples are assigned to one of these spaces, but unknown classes may be misclassified as known classes. In (b), the decision boundary is represented by a solid line, and if a sample falls within a known class region, it is identified as that class. Otherwise, it is recognized as an unknown category. The algorithm for OSR can be considered an $n+1$ classifier, where $n$ is the number of known categories in the training phase and "$1$" represents the unknown category.}
\label{consGraphs15}
\end{figure}

Another prevalent category of tasks in NPG involves analyzing and processing sequence data. One prominent of them is the signal diagnosis/abnormal detection against equipment operational status \cite{li2022research}. In industrial production scenarios, well-configured equipment must consistently maintain optimal working conditions. Thus, targeted monitoring of crucial equipment aims to identify potential operational failures. Once faults or abnormalities are detected, they need to be classified into distinct categories and corresponding measures should be taken. These works rely on analyzing the output data from sensors installed on various devices, which capture time sequence signals. DL has been extensively utilized in fault diagnosis in recent years due to its exceptional feature extraction capabilities \cite{gao2023improve}. However, its application in actual industrial environments still poses several challenges. Most current DL approaches rely on the closed-set assumption, which assumes that the test data comes from a priori known classes. Consequently, unknown classes are often hastily assigned to known classes with overconfidence. This issue becomes particularly apparent when dealing with high-dimensional feature spaces or intricate features, such as incomplete class labels and imbalanced positive/negative samples. To address these issues, DL-based recognition algorithms known as OSR have emerged (Figure~\ref{consGraphs15}). The OSR algorithm aims to identify unknown faults that have not been encountered in the training set while ensuring the accuracy of known fault diagnosis, referred to as Open-Set Fault Diagnosis (OSFD) \cite{yu2021deep}. Additionally, approaches based on the open set assumption address the limitation of traditional outlier/anomaly detection methods that are unable to reconize known classes due to their reliance on the closed set assumption. 

% This case study focuses on the fact that most current DL approaches rely on the closed-set assumption, which assumes that the test data comes from a priori known classes. In other words, these methods assume that the training set contains all data categories that occur in the environment in which the model is deployed. This limitation arises primarily due to the inability of traditional neural network classifiers to establish more precise decision boundaries, thereby struggling to effectively distinguish unknown classes. Consequently, unknown classes are often hastily assigned to known classes with overconfidence. This issue becomes particularly apparent when dealing with high-dimensional feature spaces or intricate features, such as incomplete class labels and imbalanced positive/negative samples. 

%As previously noted, the operating data in industrial production possesses extremely imbalanced positive/negative samples: most of the time, the equipment operates in a regular or predetermined mode, resulting in data that naturally belongs to the same known class or classes. However, as the running time progresses or working conditions and the external environment change, data with distinct characteristics and distributions are generated, belonging to unknown classes. These unknown classes may represent new normal modes or faults that require detection, but regardless of the scenario, the volume of such data is significantly smaller than the normal/default mode.

\begin{figure}[t]%[htbp]
\centering
\includegraphics[width=0.9\textwidth]{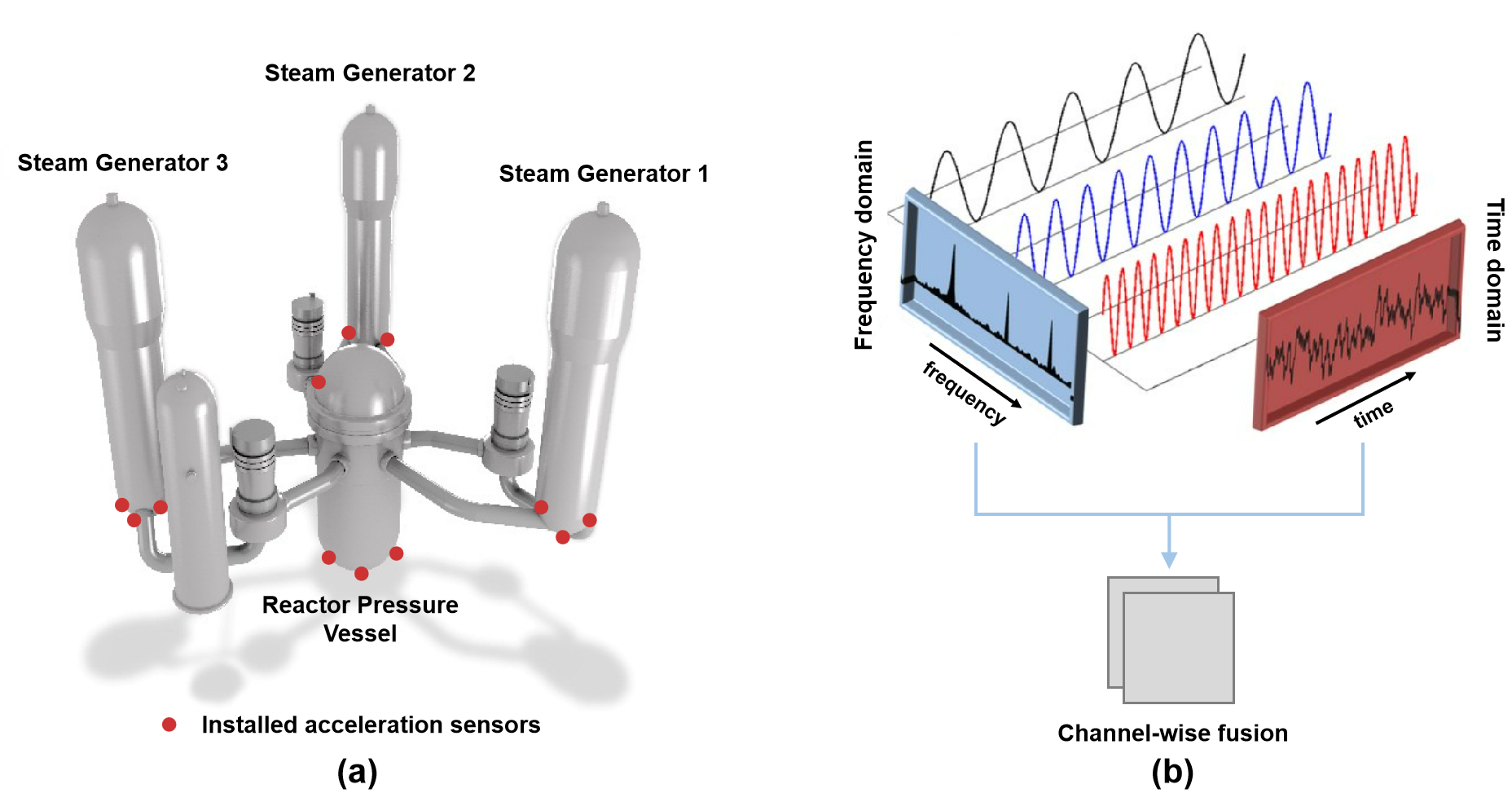}
\caption{(a) The primary circuit of the nuclear power plant with installed acceleration sensors (modified from~\cite{app10134434}) and (b) the pre-processing with multichannel time-frequency fusion (modified from~\cite{8294127}).}
\label{fig:sequence_data}
\end{figure}

In the NPG scenario, many sensors, e.g., temperature sensors, pressure sensors, and acceleration sensors, are deployed in nuclear power plants' primary and secondary circuits (Figure~\ref{fig:sequence_data}(a)). Taking the Primary Circuit of the Nuclear Reactor (PCNR) as an example, the acceleration sensors installed on the pressure vessel play a crucial role in assisting maintenance personnel in determining the working state of the PCNR by collecting vibration signals and performing analysis and processing. This case study proposed a novel framework named Open-Set Signal Recognition (OSSR). In OSSR, the sensor signal data will first be pre-processed with time-frequency fusion \cite{tao2023unsupervised}. Then, the Deep Variational Encoder-Classifier (DVEC), which combines a variational Bayesian network \cite{liu2020probabilistic} and a supervised learning paradigm, will enhance the features of the input data in the hidden layers. OSSR thus gains the ability to construct accurate decision boundaries. Eventually, an open-set discriminator block is further designed for DNN, incorporating the Extreme Value Theory (EVT) and Shannon's entropy theory. This approach effectively mitigates the risk associated with open set faults in OSFD caused by the lack of prior knowledge of unknown classes. In addition, it enhances the reliability of detecting unknown classes.

% \begin{figure}[t]
%   \centering
%   \subfloat[\label{consGraphs16}]{
%     \includegraphics[width=0.49\textwidth]{img/sensor.png}
%   } 
%   \subfloat[\label{consGraphs17}]
%   {
%     \includegraphics[width=0.49\textwidth]{img/FFT.png}
%   }  
%   \caption{(a) The primary circuit of the nuclear power plant with installed acceleration sensors. (b) Multichannel time-frequency fusion pre-processing.}
%   \label{fig:data_charac}
% \end{figure}

% \begin{figure}[t]%[htbp]
% \centering
% \includegraphics[width=0.44\textwidth]{pic/Fig12_3.png}
% \caption{The primary circuit of the nuclear power plant with installed acceleration sensors.}
% \label{consGraphs16}
% \end{figure}

% \begin{figure}[htbp]%[htbp]
% \centering
% \includegraphics[width=0.43\textwidth]{pic/Fig13_3.png}
% \caption{Multichannel time-frequency fusion pre-processing.}
% \label{consGraphs17}
% \end{figure}

\subsubsection{Pre-processing of the time sequence signal data}
To begin discussing the DVEC, which serves as the core component of OSSR, it is essential to pre-process the raw data input to the model. This is mainly to alleviate the learning burden of the model through preliminary signal analysis. According to Fourier's theory, any continuously measured time sequence signal can be expressed as a combination of sinusoidal signals with varying frequencies. From this, the Fourier transform method has become a widely used technique in signal processing. By converting a signal obtained from a measurement from the time domain to the frequency domain, the signal can be efficiently separated based on its different frequency components. In practical applications, the Fast Fourier Transform (FFT) is commonly employed to calculate the sequence in the time domain as a frequency domain value or frequency component \cite{eleftheriadis2022energy}. It reduces computational costs and enables comprehensive spectrum analysis of all data points within high-frequency resolution samples. 

However, most of the existing methods typically utilize only individual signal features in either the time or frequency domain as inputs for the diagnostic model. This led to their imperfections in capturing the wide range of diverse and variable operating statuses, making it challenging to meet intricate industrial requirements. A more viable approach involves performing a time-frequency analysis to derive time-frequency patterns that correspond to different operating statuses. The OSSR framework tackles this issue by employing FFT to convert the raw sensor data, i.e. time domain sequence signals, into frequency domain sequence signals \cite{cao2021multi}. Subsequently, it fuses the data from both time and frequency domains using a multichannel approach, treating them as inputs for the network model (Figure~\ref{fig:sequence_data}(b)). This preprocessing technique allows for the comprehensive utilization of the signal's time and frequency domain information. It plays a crucial role in enabling the transfer of the OSR model from non-sequential data tasks to sequential signal tasks in OSFD problems. Additionally, it overcomes the limitations of existing methods in extracting robust features when applied to signal datasets.

\subsubsection{Deep Variational Encoder-Classifier}
Traditional DL models rely on labeled data to learn decision boundaries, limiting their ability to classify only known classes. To accurately recognize the unknown classes, a robust model must learn features from the data and thus establish decision boundaries that can distinguish these classes. One approach to achieve this is through the use of Variational Autoencoder (VAE), an advanced probabilistic distribution modeling technique that combines variational Bayesian inference with a decoder-encoder structured Neural Network (NN) to extract the latent feature vector of the data \cite{girin2020dynamical}. The fundamental concept behind VAE is that any data distribution can be viewed as a mixture of multiple Gaussian distributions. VAE leverages this concept to approximate the posterior $q_{\phi}(z|x)$ to approach the true prior $p_{\theta}(z)$ by modeling the log-likelihood $log p_{\theta}(x)$ of the input data. This allows VAE to detect unknown classes efficiently.

\begin{figure*}[t]%[htbp]
\centering
\includegraphics[width=0.98\textwidth]{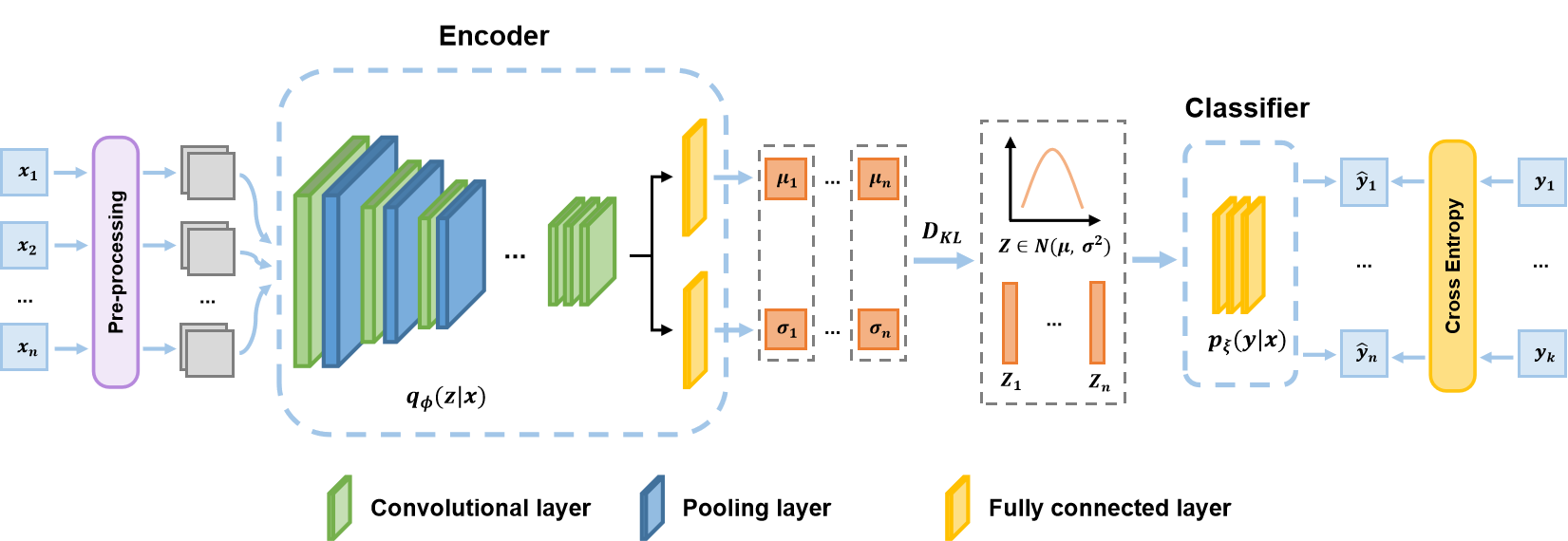}
\caption{The workflow of DVEC: before entering the NN model, the raw time-domain signal data is pre-processed using FFT to convert it to the frequency domain. Then, a multichannel fusion of time and frequency domains is performed at the channel level to enhance discriminative power and reduce information loss, resulting in improved fault diagnosis. In DVEC, the encoder network with parameter $\phi$ learns the mapping $q_{\phi}(z|x)$ from the input $x_{i}$ to the low-dimensional latent feature $z_{i}$  using a probabilistic ladder structure in each layer. The mapping is shown to be mixed Gaussian distributed, with two fully connected layers learning the expectation $\mu$ and variance $\sigma^{2}$  and constraining the mixed Gaussian distribution $N(\mu, \sigma^{2})$ over $z$ by KL divergence. Subsequently, the classifier learns the mapping $p_{\xi}(y|z)$ from the latent feature $z_{i}$ to the category label $y_{i}$ and outputs $\hat{y}_{i}$, where $y_{i}$ belongs to the dataset $\{{y_{i}}\}^k_{i=1}$ and $k$ is the number of categories of known classes in the training phase. Finally, the classification of known classes and the correction of latent features $z$ are performed using the classification loss function.}
\label{consGraphs18}
\end{figure*}

However, both VAE and its improvement, Deep Variational Encoder Classifier (DVAEC), suffer from the issue of inadequate ability to classify known classes based on latent features. VAE is insufficient in providing discriminative features for known classes and unknown detection since it does not introduce labeling information during training. To address this limitation, the newly proposed DVEC enhances the features of the input data in the hidden layer by introducing a supervised learning paradigm in VAE while using a probabilistic ladder structure to extract high-level abstract features. DVEC comprises an encoder, a classifier, and a loss function $L(\phi, \xi; x, y)$ (Figure~\ref{consGraphs18}). The encoder is a NN with parameters $\phi$, which takes the sample $x$ as input and outputs the latent feature representation $z$. The classifier is another NN with parameter $\xi$, which takes the representation $z$ as input and learns the mapping of $z$ to the category label $y$. The main difference between DVEC and VAE is that DVEC adopts a linear classifier $p_{\xi}(y|z)$ to maximize the lower bound of $p(y)$ instead of $p(x|z)$. This is feasible because the input $x$ is strongly associated with the label $y$, and the process of reconstructing $x$ by $z$ can be replaced by learning the mapping of $z$ to label $y$. During the training phase of DVEC, features subject to variational constraints can be learned by optimizing the loss function shown in Eq.~\eqref{DVECLoss} to obtain more accurate decision bounds:
\begin{equation}\label{DVECLoss}
\begin{gathered}
L(\theta, \xi ; x, y)=\mathbb{E}_{q_\phi(z \mid x)}\left[\log p_{\xi}(y\mid z)\right]-\lambda(x) D_{KL}\left(q_\phi(z\mid x) \| p_\theta(z)\right).
\end{gathered}
\end{equation}

The second term on the right-hand side, which is essentially identical to that in VAE, enforces a Gaussian distribution of the features in the latent feature space according to the corresponding labels through KL divergence. This effectively reduces intra-class dispersion in the feature space. In contrast, the first term represents the classification error of the classifier when recognizing known class samples, rather than the reconstruction error in VAE. This flexible substitution enables the network to prioritize known classifications subjecting to variational constraints. The weight parameter $\lambda(x)$ is also introduced to the equation (Eq.~\eqref{AW}). The over-constraint of the KL term is avoided by setting $\lambda(x)$ as a weighting factor between 0 and 1. The literature \cite{mundt2019open} indicates that the weight parameter can help alleviate the posterior collapse problem in Bayesian models:
\begin{equation}\label{AW}
\begin{gathered}
\lambda(x)= \begin{cases}\max \left(\omega(x), \omega_0\right), & \text { if } y=\hat{y}_i ,\\ 0, & \text { or } y \neq \hat{y}_i.\end{cases}
\end{gathered}
\end{equation}

\begin{table}[t]
\begin{center}
\caption{Labels of the Primary Circuit of the Nuclear Reactor (PCNR) Dataset.}
\label{labelsofthepublicdataset} \scalebox{0.95}{
\begin{tabular}{ccc|ccc}
\hline\hline
\textbf{No.} & \textbf{Categories}                                              & \textbf{Implicatio}                                                                                   & \textbf{No.} & \textbf{Categories} & \textbf{Implicatio}                                                                                         \\ \hline
1            & Noise                                                            & Signals with high ground noise                                                                        & 2            & Spike               & Occasional impulse spike signals                                                                            \\ \hline
3            & Jam                                                              & \begin{tabular}[c]{@{}c@{}}Unidentified jamming \\ signal in the reactor\end{tabular}                 & 4            & Impact              & \begin{tabular}[c]{@{}c@{}}Impact wave signal generated by \\ the impact of metal foreign body\end{tabular} \\ \hline
5            & \begin{tabular}[c]{@{}c@{}}Hydraulic \\ fluctuation\end{tabular} & \begin{tabular}[c]{@{}c@{}}Signals triggered by hydraulic \\ fluctuations of the circuit\end{tabular} & 6            & Self-check          & \begin{tabular}[c]{@{}c@{}}Signals generated during the self-\\ check of the monitoring system\end{tabular} \\
\hline\hline
\end{tabular}}
\end{center}
\end{table}

\subsubsection{Open Set Discriminator based on EVT and Entropy}
To further improve the discriminative capability of OSSR for unknown classes, an unknown class discriminator based on the open set assumption was specifically designed based on DVEC as a way to improve the contribution of sample latent features to discrimination. Specifically, two different strategies were employed to build the discriminator, EVT and Shannon's entropy. EVT is widely used in the current OSR because it effectively fits the tail distribution of the network's output. In OSSR, a score is calculated based on the distance between the latent features and the high-density region, and then the EVT model is fitted using the Weibull distribution. During the validation phase, the statistical outlier probability threshold of the Cumulative Distribution Function (CDF) is determined based on the distribution of the latent features of the known classes (Eq.~\eqref{CDF}). Samples from unknown classes in the testing phase are determined to be rejected or not based on this threshold.
\begin{equation}\label{CDF}
\begin{gathered}
p_{k, j}=1- \exp{\left(-\left(\frac{\left\|d_{k, j}-\tau_k\right\|}{\gamma_k}\right)^{\kappa_k}\right)},
\end{gathered}
\end{equation}
\noindent where $d_{k, j}$ denotes the distance from the validation set samples to the class center of the known class $k$. $\tau_k$, $\kappa_k$, and $\gamma_k$ are the parameters of Weibull distribution model $\rho_{k}(\tau_{k}, \kappa_{k}, \gamma_{k})$.

However, relying solely on EVT to recognize unknown classes can result in overly simplistic discriminative criteria, leading to weak model generalization and reduced performance when faced with difficult-to-fit feature distributions. Additionally, fitting the feature distribution requires a sufficient number of samples from known classes to ensure reliability. To address these limitations, the threshold selection method based on Shannon's entropy (Eq.~\eqref{Entropy}) leverages the mathematical measure of uncertain probability events:
\begin{equation}\label{Entropy}
\begin{gathered}
H(X)=-\sum_{x\in X} p(x)\log p(x),
\end{gathered}
\end{equation}
\noindent where $p(x)$ denotes the probability that the value of the variable $X$ is $x$. The larger the entropy $H(X)$, the greater the uncertainty of $X$.

\begin{figure}[b]%[htbp]
\centering
\includegraphics[width=0.97\textwidth]{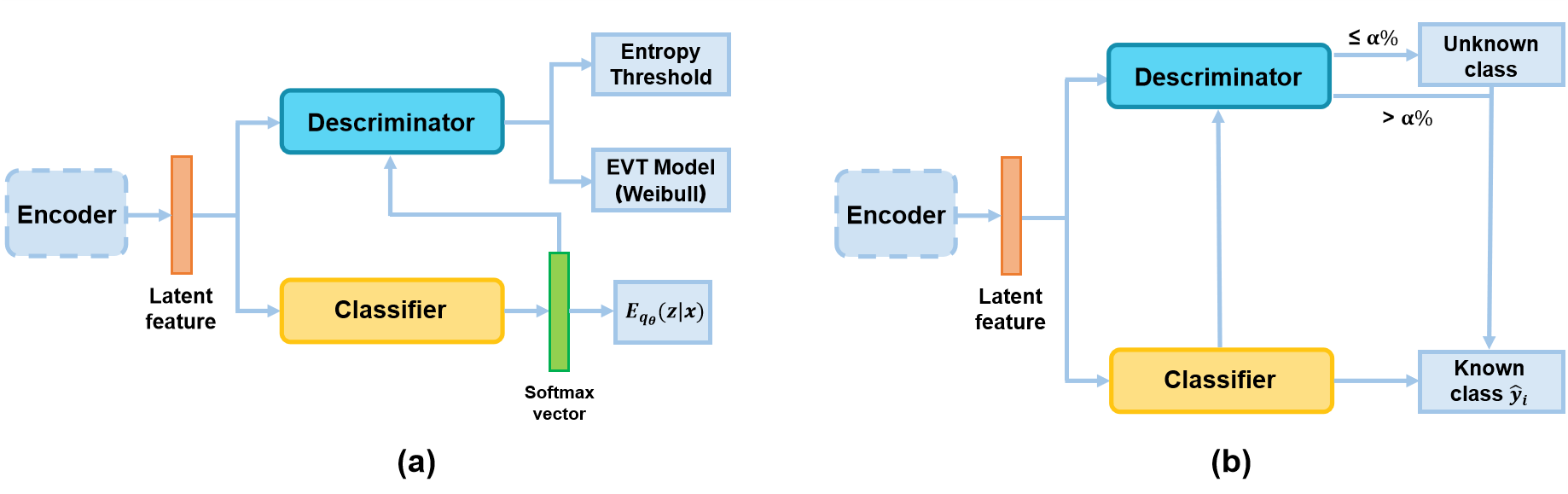}
\caption{The workflow of the open set discriminator based on (a) EVT and (b) entropy.}
\label{consGraphs19}
\end{figure}

% \begin{figure}[t]
%   \centering
%   \subfloat[Validation Phase\label{consGraphs19-a}]{
%     \includegraphics[width=0.45\textwidth]{img/flowcase2-a.png}
%   } 
%   \subfloat[Testing Phase\label{consGraphs19-b}]
%   {
%     \includegraphics[width=0.45\textwidth]{img/flowcase2-b.png}
%   }  
%   \caption{The workflow of the open set discriminator based on Extreme Value Theory (EVT) and entropy.}
%   \label{consGraphs19}
% \end{figure}

Specifically, the entropy value of the softmax probability vector of the final output can be calculated to represent the uncertainty of the network output. Samples of unknown classes can be accurately identified by selecting an appropriate uncertainty threshold, similar to the CDF threshold selection rule. Figure~\ref{consGraphs19} depicts the workflow of the open set discriminator, which can be segregated into two distinct phases, namely validation and testing:

\begin{itemize}
\item \textbf{Validation phase:} as shown in Figure~\ref{consGraphs18} and \ref{consGraphs19}(a), the trained encoder outputs the latent features of all the validation set samples, which are subsequently utilized as inputs to both the classifier and discriminator. The softmax vector produced by the classifier's final layer is also fed into the discriminator. The discriminator leverages EVT to fit the Weibull distribution model to the latent features of each known class. Furthermore, the entropy value is computed for the softmax vector of each known class, thereby generating a list of entropies that corresponds to each known class.

\item \textbf{Testing phase:} during this phase (Figure~\ref{consGraphs19}(b)), the encoder provides the latent features of the test dataset. The classifier initially predicts a known class $\hat{y}_i$. The discriminator computes the CDF of the extracted latent feature using the Weibull distribution model of class $\hat{y}_i$ to estimate the outlier probability. The discriminator then sets the outlier probability of the $\alpha\%$ highest values as the dynamic threshold based on the acceptance rate $\alpha\%$ set by the hyperparameter. It determines whether the current outlier probability of the latent feature in this Weibull distribution model is greater than the threshold. If it is, the sample is classified as an unknown class. The softmax vector generated by the classifier is fed into the discriminator, and the list of entropies of this class calculated in the validation stage is also set. The entropy of the $\alpha\%$ highest values is selected as the dynamic threshold. Similarly, the entropy of this softmax is recognized as an unknown class if it exceeds the threshold value. 
\end{itemize}  

\subsubsection{Validation and Evaluation}
\begin{figure}[t]%[htbp]
\centering
\includegraphics[width=0.92\textwidth]{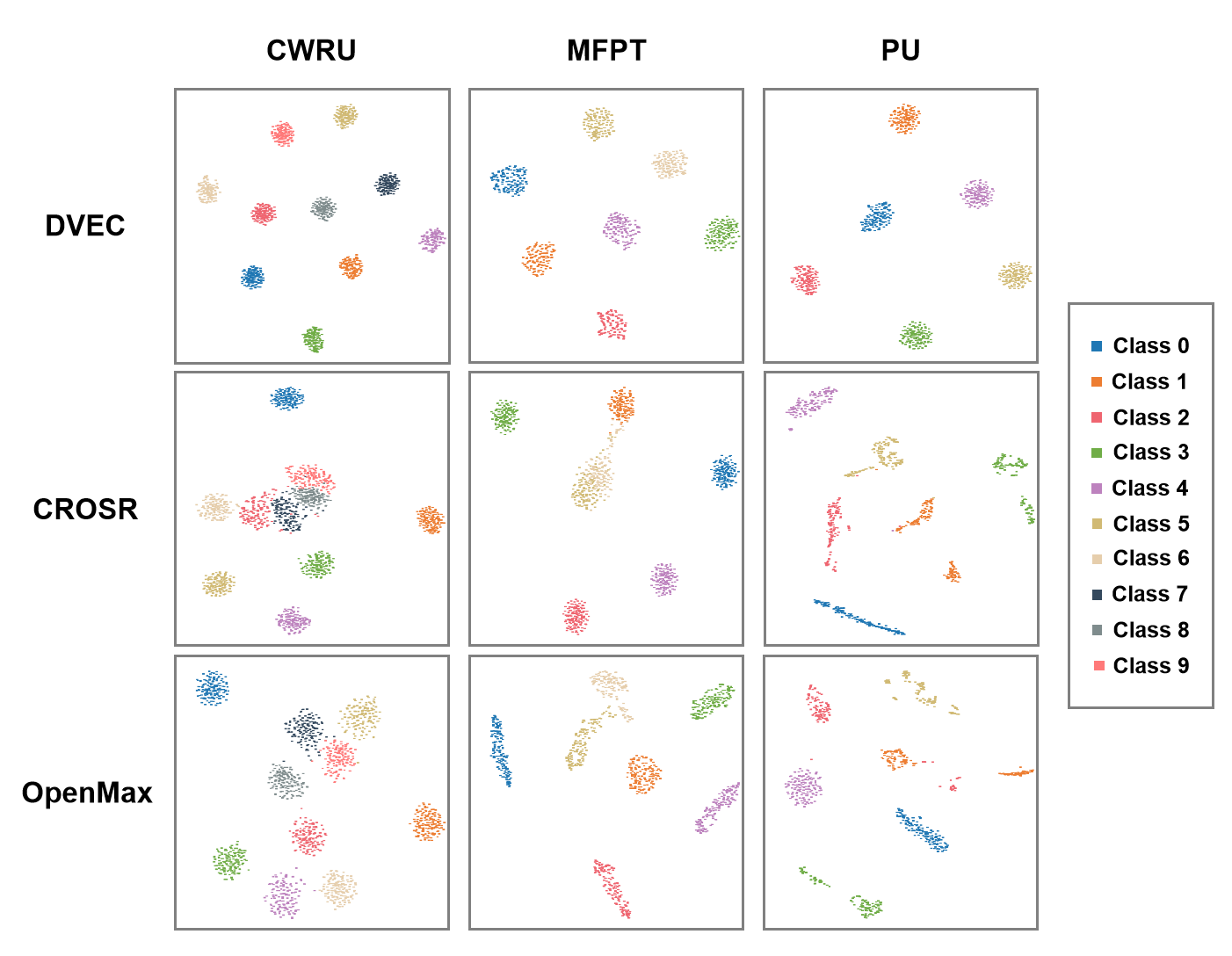}
\caption{The visualization of feature extraction results of DVEC, OpenMax and CROSR on CWRU, MFPT, and PU datasets: DVEC disperses the latent feature regions of different categories and ensures a high degree of aggregation within categories; in contrast, the generative model represented by CROSR performs feature extraction with no clear distinction and overlap in the feature distribution of each category, which indicates the limited use of these methods for category distinction; for OpenMax, since no constraints are placed on the feature distribution, the extracted features not only have no specific distribution but also overlap between categories.}
\label{consGraphs20}
\end{figure}

Eventually, the newly proposed approaches have demonstrated outstanding performance in various tests and real-world applications, including the extraction of hidden features from publicly available fault diagnosis datasets such as CWRU (the Bearing Dataset from Case Western Reserve University), MFPT (the Bearing Fault Dataset from the society of Machinery Failure Prevention Technology), and PU (the Bearing Dataset from Paderborn University)~\cite{Lessmeier_Kimotho_Zimmer_Sextro_2016}, as well as open set fault diagnosis on the PCNR dataset. Specifically, DVEC has outperformed other methods in key items such as the well-distributed features and inter-category differentiation for the task of feature extraction. This superiority is evident from the visualized results using the t-SNE (t-Distributed Stochastic Neighbor Embedding) method, as shown in Figure~\ref{consGraphs20}, where DVEC surpasses the baseline approaches such as CROSR and OpenMax.

\begin{table}[t]
\begin{center}
\caption{Open Set Fault Diagnosis tasks on the Primary Circuit of the Nuclear Reactor (PCNR) dataset.}
\label{missions} \scalebox{0.95}{
\begin{tabular}{cccc}
\hline\hline
\rule{0pt}{12pt}
\textbf{Msn.} & \textbf{Known classes} & \textbf{Unknown classes} & \textbf{Openness}\\ 
\cline{1-4}
\rule{0pt}{16pt}
1 & 2 & 0, 1, 3, 4, 5 & 0.5918\\
\rule{0pt}{16pt}
2 & 2, 3 &  0, 1, 4, 5 & 0.4226\\
\rule{0pt}{16pt}
3 & 0, 1, 2 & 3, 4, 5 & 0.2929\\
\rule{0pt}{16pt}
4 & 2, 3, 4, 5 & 0, 1 & 0.1835\\
\rule{0pt}{16pt}
5 & 0, 1, 2, 4, 5 & 3 & 0.0871\\
\hline\hline
\end{tabular}}
\end{center}
\end{table}

Furthermore, to validate the recognition performance and generalization of the EVT and entropy-based discriminators, they were integrated with DVEC, OpenMax, and CROSR and applied to the OSR task of the PCNR dataset. The PCNR dataset is comprised of vibration signals induced by local impacts in the PCNR, which were collected by sensors during the actual operation of the reactor, and subsequently constructed as a corresponding dataset. The vibration signal from each measurement point was sampled at a frequency of 50,000 samples/second, with each sample consisting of 5,000 sampling points. After expert classification, these signals were classified into six categories, with the meanings of each category and corresponding labels (0 to 5) shown in Table~\ref{labelsofthepublicdataset}.

% Additionally, it is believed that various OSR tasks exhibit distinct characteristics, in particular, the ratio change between the known and unknown classes can considerably impact the outcomes of OSFD. This aspect can be articulated by adopting the notion of openness as formulated by Scheirer et al. Meanwhile,  as the category set utilized for OSR testing is identical to the category set for identification, a simplified rendition can be expressed using Eq.~\eqref{openness}:

It is widely accepted that different OSR scenarios possess distinct characteristics, with the ratio shift between known and unknown classes playing a significant role in determining OSR performance. This notion can be better understood by embracing the concept of openness, as defined by Scheirer et al. Additionally, since the category set employed for OSR testing is identical to that used for recognition, a simplified version can be represented using Eq.~\eqref{openness}:
\begin{equation}\label{openness}
\begin{gathered}
openness=1-\sqrt{\frac{N_{training}}{N_{testing}}},
\end{gathered}
\end{equation}
\noindent where $N_{training}$ denotes the number of classes in the training set, i.e., the set size of known classes, and $N_{testing}$ denotes the number of classes in the test set, i.e., the sum of the number of known classes and unknown classes. 

To assess the efficacy of the novel approach in various open set scenarios, a series of tasks featuring different levels of openness were conducted using the PCNR dataset. As depicted in Table~\ref{missions}, each job involved selecting a subset of classes from the dataset to form the known class set, while the remaining classes were considered unknown and used solely during the testing phase. This approach also accounted for the possibility of an extreme scenario where only one known class was present. This issue is not uncommon in practical applications, such as when a new type of device is applied into production and many of its characteristics are still unknown. The results of the experiments are presented in Table~\ref{PCNRresults}, and it is evident that the new method outperforms several other methods that were compared, thereby achieving the most superior results. 

%which demonstrates that OSR of the PCNR dataset obtained from a practical industrial environment, remained a challenging task for many other approaches, owing to the high dimensionality of the data samples and significant levels of noise.

\begin{table}[t]
\centering
\begin{center}
\caption{Accuracy $A_{0}$ of tasks performed by DVEC, OpenMax, and CROSR with discriminators based on EVT and Entropy on the PCNR dataset: DVEC exhibits the best performance; CROSR based on generative methods possesses relatively higher robustness towards task openness as it learns the reconstruction of latent features as data; the openmax layer vectors defined by OpenMax are not robust enough to be considered as reliable features.}
\label{PCNRresults} \scalebox{0.95}{
\begin{tabular}{cc|ccccc}
\hline\hline
\multirow{2}{*}{\textbf{Methods}}             & \multirow{2}{*}{\textbf{Descriminator}} & \multicolumn{5}{c}{\textbf{Mission}}                                                    \\ \cline{3-7} 
                                              &                                         & \textbf{1}      & \textbf{2}      & \textbf{3}      & \textbf{4}      & \textbf{5}      \\ \hline
\multicolumn{1}{c|}{\multirow{2}{*}{DVEC}}    & EVT                                     & \textbf{0.9555} & \textbf{0.9019} & 0.8061          & 0.9840          & 0.9609          \\
\multicolumn{1}{c|}{}                         & Entropy                                 & /               & 0.8719          & \textbf{0.9804} & \textbf{0.9875} & \textbf{0.9769} \\ \hline
\multicolumn{1}{c|}{\multirow{2}{*}{OpenMax}} & EVT                                     & 0.3950          & 0.3559          & 0.4715          & 0.7117          & 0.8861          \\
\multicolumn{1}{c|}{}                         & Entropy                                 & /               & 0.6317          & 0.6940          & 0.9751          & 0.9395          \\ \hline
\multicolumn{1}{c|}{\multirow{2}{*}{CROSR}}   & EVT                                     & 0.4377          & 0.5765          & 0.8523          & 0.7135          & 0.8416          \\
\multicolumn{1}{c|}{}                         & Entropy                                 & /               & 0.8238          & 0.7829          & 0.8274          & 0.7740          \\ \hline\hline
\end{tabular}}
\end{center}
\end{table}

What can also be seen is that as the task openness decreases, meaning the proportion of known classes increases, the entropy-based discriminator would gradually outperform the EVT version. This is due to the increase in reliability of the computed entropy as the number of prediction vector elements of the network output increases. Conversely, with fewer known classes, the network is more likely to learn a more discriminative posterior distribution space for the latent features of known classes, making the distribution deviation probabilities derived from the EVT distribution model more reliable. The OSSR, consisting of DVEC and this pair of discriminators, thus achieves exceptional performance across all tasks in the setup. Notably, the results of this experiment also revealed that in some actual applications, the discriminators may provide two distinct discriminations for a single sample. The final decision can be made based on the given task's openness. For instance, the entropy-based discriminator may be deemed more reliable in low openness scenarios.

\section{Discussion and Conclusion}
\label{future}
Despite significant advancements in AI technology applications in industrial production, there are still challenges in the theoretical research and application exploration of related methods, particularly in intricate scenarios such as the full life-cycle intelligence of NPG. One of the main challenges is that the mainstream development of AI methods, represented by DL, is currently focused on the fields of image processing and natural language processing, whereas the data structure generated by nuclear power generation scenarios is considerably different from the data structure of these fields. Furthermore, a significant portion of the data in nuclear power plants is either difficult to access or, worse, to annotate for security reasons, posing additional challenges and difficulties in applying AI directly to nuclear power generation. Additionally, given the critical nature of issues related to nuclear power generation, such as safety and environmental protection, high performance is required in terms of accuracy, fault tolerance, interpretability, and real-time performance of algorithms.

In this paper, the potential role of deep learning technologies with finite samples in the full life-cycle intelligence of NPG has been explored and two case studies were presented. The main conclusions are obtained as follows:
\begin{itemize}
\item After conducting an in-depth analysis of the application scenarios where AI can be employed in the full life-cycle of NPG, this paper identifies the challenges faced by DL methods in terms of data structure and security concerns, including long-tailed class distribution, sample imbalance, and domain shift. Based on these, we review the potential of deep learning with finite samples, including small-sample learning, few-shot learning, zero-shot learning, and open-set recognition.
\item In the case of automatic recognition of zirconium alloy metallography, the available training samples for building up DL models are significantly smaller than in other scenarios, while the labeling cost is considerably higher. To address this issue, data augmentation and transfer learning techniques have been used to assist in the model's training. The final result confirms that the proposed approach of small-sample learning gives the neural network the ability to train efficiently on the basis of a small number of samples and integrate it with other prior knowledge, similar to the "lift-and-shoot" of humans.
\item The case study on open-set recognition-based signal diagnosis constructs a more refined decision boundary than previous closed-set-based assumptions by delving deeper into the latent features of sequential signals (vibrations) obtained from accelerometers. This approach also enables the recognition of many unknown classes of failures that are critical to safety during reactor operation, which were previously overlooked, significantly improving the system's safety.
\end{itemize}

Overall, in contrast to earlier utilizations that primarily associated AI with big data, intelligence in industrial scenarios must contend with massive amounts of operational data that are highly imbalanced in terms of positive (normal) and negative (abnormal) samples due to safety and other considerations. This imbalance greatly complicates the learning process for network models. Additionally, relying solely on data-driven training methods has become increasingly unsophisticated and requires improvement. However, by utilizing mathematical tools such as Fourier transform to extract and analyze few anomalous samples more efficiently, and then constructing samples' feature space in a more refined manner, deep learning with finite samples can effectively address these challenges. This paper provides a deeper understanding of the role of deep learning with finite samples in industrial intelligence and proposes specific methods to enhance the development and utilization of clean energy in a safer and more efficient manner, especially in complex scenarios such as the full life-cycle intelligence of NPG. As research on industrial intelligence in NPG progresses, the range of application scenarios will continue to be expanded and more specific methods will be proposed to enhance the development and utilization of this clean energy in a safer and more efficient manner.

\section*{Declaration of Competing Interest}
The authors declare that they have no known competing financial interests or personal relationships that could have appeared to influence the work reported in this paper.

\section*{Acknowledgement}
This work is supported by the National Science Foundation of China under Grant 62106161, the Key Program of National Science Foundation of China under Grant 61836006, the Key R\&D Program of Sichuan Province under Grant 2022YFN0017, and "The Fundamental Research Funds for the Central Universities" under Grant 2022SCU12072.

\printcredits
\bibliographystyle{elsarticle-num}
\bibliography{cas-refs}

% % Biography
% \bio{}
% % Here goes the biography details.
% \endbio

% \bio{pic1}
% % Here goes the biography details.
% \endbio

\end{document}